\documentclass[12pt]{article}

\usepackage{graphicx}
\usepackage{amssymb}
\usepackage{amsthm}
\usepackage{booktabs}
\usepackage{rotating}
\usepackage{multirow}
\usepackage{eurosym}
\usepackage{amsfonts}
\usepackage{amsmath}
\usepackage{umoline}
\usepackage{caption}
\usepackage{subcaption}
\usepackage{color}
\usepackage{url}
\usepackage{authblk}
\usepackage[english]{babel}

\usepackage{pbox}
\usepackage{footnote}
\usepackage{multirow}
\usepackage{changepage}
\usepackage{tablefootnote}

\usepackage[margin = 2cm]{geometry}

\usepackage{hyperref}

\usepackage{float}
\usepackage{multirow}
\usepackage{setspace}
\usepackage{natbib}
\usepackage{booktabs}
\usepackage{hyperref}
\usepackage{tabularx}
\usepackage{ccicons}

\setcitestyle{authoryear,open={(},close={)}}
\begin{document}

\title{Deep residential representations: Using unsupervised learning to unlock elevation data for geo-demographic prediction \footnote{\scriptsize NOTICE: This is a preprint of a published work. Changes resulting from the publishing process, such as editing, corrections, structural formatting, and other quality control mechanisms may not be reflected in this version of the document. Please cite this work as follows: Stevenson, M., Mues, C., \& Bravo, C. (2022). Deep residential representations: Using unsupervised learning to unlock elevation data for geo-demographic prediction. ISPRS Journal of Photogrammetry and Remote Sensing, 187, pp.378-392 DOI: \protect\url{https://doi.org/10.1016/j.isprsjprs.2022.03.015}. This work is made available under a \href{https://creativecommons.org/licenses/by-nc-nd/2.0/}{Creative Commons BY-NC-ND license}. \ccbyncnd}}

\author[1,2]{Matthew Stevenson}
\author[1,2]{Christophe Mues}
\author[3]{Cristi\'{a}n Bravo}

\affil[1]{Department of Decision Analytics and Risk, Southampton Business School, University of Southampton, University Road, SO17 1BJ, United Kingdom.}

\affil[2]{Centre for Operational Research, Management Sciences and Information Systems (CORMSIS), University of Southampton, University Road, SO17 1BJ, United Kingdom.}

\affil[3]{Department of Statistical and Actuarial Sciences, The University of Western Ontario, 1151 Richmond Street, London, Ontario, N6A 5B7, Canada.}

\date{}

\maketitle

\begin{abstract}
LiDAR (short for ``Light Detection And Ranging'' or ``Laser Imaging, Detection, And Ranging") technology can be used to provide detailed three-dimensional elevation maps of urban and rural landscapes. The geographically granular and open-source nature of this data lends itself to an array of societal, organisational and business applications where geo-demographic type data is utilised. However, the complexity involved in processing this multi-dimensional data in raw form has thus far restricted its practical adoption. This paper proposes a series of convenient task-agnostic tile elevation embeddings to address this challenge, using recent advances from unsupervised Deep Learning. We test the potential of our embeddings by predicting seven English indices of deprivation (2019) for small geographies in the Greater London area. These indices cover a range of socio-economic outcomes and serve as a proxy for a wide variety of potential downstream tasks to which the embeddings can be applied. We consider the suitability of this data not just on its own but also as an auxiliary source of data in combination with demographic features, thus providing a realistic use case for the embeddings. Having trialled various model/embedding configurations, we find that our best performing embeddings lead to Root-Mean-Squared-Error (RMSE) improvements of up to 21\% over using standard demographic features alone. We also demonstrate how our embedding pipeline, using Deep Learning combined with K-means clustering, produces coherent tile segments which allow the latent embedding features to be interpreted.
\end{abstract}


%%\linenumbers
\section{Introduction}\label{intro}

In recent years, there has been a growing trend in the application of remote sensing to predict or explain socio-economic phenomena. The appeal of this data source is that it provides a relatively inexpensive means of capturing vast amounts of information at a geographically fine-grained level and at regular time intervals. Perhaps the most natural fit for remote sensing with such applications is in the developing world, where remote sensing can support or replace existing data sources without reliable structured census data. For example, satellite data has been successfully applied to poverty prediction \citep{block2017unsupervised, jean2016combining}. More recently, satellite imaging and street view imagery have also been used in the developed world to improve house price estimation \citep{law2019take}, detect abandoned houses \citep{zou2021abandoned} and predict deprivation indices \citep{suel2021multimodal}. Instead of acting as a substitute, socio-economic data and satellite imagery have also been used in combination in order to classify urban scenery \citep{su2021urban}. However, while forms of data such as satellite and street imagery are predictive of various outcomes, they also present challenges. Firstly, the data is often proprietary or available under an academic licence only, restricting its practical adoption. There is also an ethical question as to the intrusiveness of these types of images when being used to make sensitive decisions, for example, to assess a person's creditworthiness. 

\begin{figure}[!htb]
\centering
\scalebox{1}{
  \includegraphics[width=0.8\linewidth]{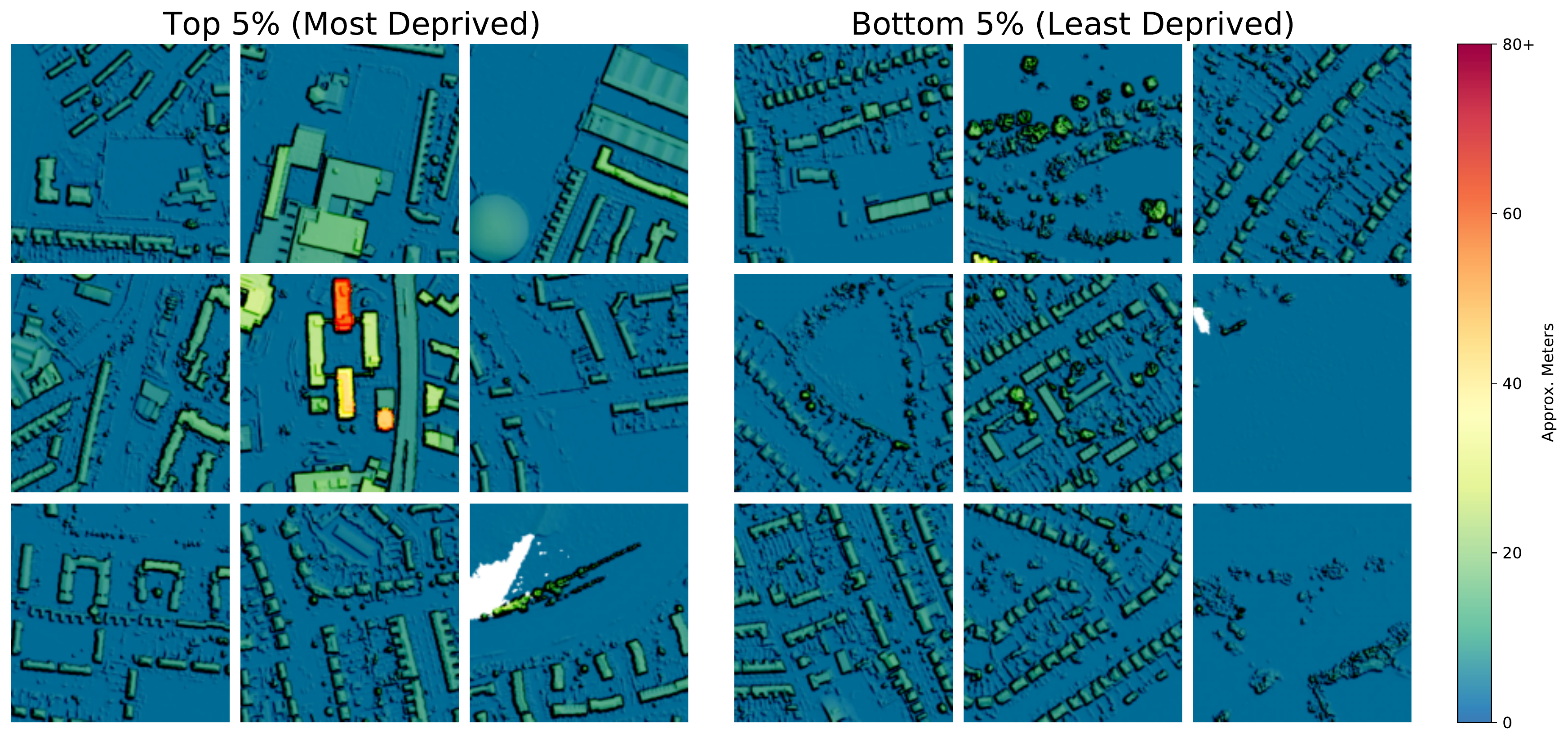}}
  \caption{Example of LiDAR elevation tiles}
  \label{fig:ExampleElevations}
\end{figure}

LiDAR, short for ``Light Detection And Ranging'' or ``Laser Imaging, Detection, And Ranging", is an alternative form of remote sensing that can measure the vertical structures of the land below and provides an open-source high-resolution alternative to satellite imagery (see Figure \ref{fig:ExampleElevations}). Its capacity to capture rich imagery has led to its adoption for a range of purposes in the environmental sciences \citep{shaker2019automatic,pan2020land,hamraz2019deep,cao2019estimating, zhou2020lidar,wang2020extraction} and archaeology \citep{de2005using,sittler2007potential}. Several studies have also used LiDAR to derive handcrafted metrics that capture features from the urban environment to predict or explain socio-economic phenomena \citep{lu2011volumetric,lu2013remote,grove2014ecology,shanahan2014socio,warth2020prediction}. Such metrics may reflect environmental aspects including the levels of vegetation or attributes of the built environment such as building volume and footprint. While these methods are simple to interpret, they reduce this highly complex imagery to a single metric or set of metrics, and in the process, discard potentially valuable information that may be useful in a predictive setting. Our paper addresses this by applying advanced feature extraction methods to capture complex latent features from LiDAR imagery.

Due to LiDAR's broad environmental uses, it is commonly collected by government environmental agencies and subsequently open-sourced. For example, in the UK, LiDAR is updated and published annually under an Open Government Licence \citep[OGL;][]{ea2021lidar}, a nonrestrictive licence that allows the data to be exploited both commercially and non-commercially. Similarly, LiDAR data is made available, to varying degrees of coverage and granularity, by agencies in the USA \citep{usgs}, Canada \citep{nrc}, Australia \citep{icsm}, Ireland \citep{gsi} and the Netherlands \citep{pdok}, among others. Despite its wide availability, we argue that LiDAR data has been underutilised and could form a valuable auxiliary data source for a wide range of predictive tasks. The findings from the research by \citet{zuiderwijk2012socio} into the impediments of open data give insight as to why LiDAR has not yet been more widely adopted. They identify ten core requirements for the adoption of open data, including 1) availability and access, 2) findability, 3) usability, 4) understandability, 5) quality, 6) linking and combining data, 7) comparability and compatibility, 8) metadata, 9) interaction with the data provider, and 10) opening and uploading. While LiDAR data might meet some of these criteria, we argue that more can be done to improve its usability and ease by which it can be combined with other data sources. 

Part of the issue with the data is that it is published as a series of 3D raster tiles that require specialist knowledge and software to be processed. Even with the capacity to process the files, to use the data for a general statistical or analytics task (descriptive, predictive or prescriptive) adds a layer of complexity as, in practice, it would require the training of sizeable Deep Learning models. Furthermore, even with such capacity available, it may not seem a worthwhile exercise for practitioners when LiDAR might just form an auxiliary form of data, despite the potential predictive performance gains its use might entail. To address this, we propose a series of task-agnostic embeddings that capture higher-level features of the urban environment numerically and can be easily imported into any analytics tool. In turn, this allows LiDAR data to be practically applied to a wide array of potential tasks where geo-demographic data is of use. Furthermore, we make this data adoptable as census type data, which is both geographically fine-grained and convenient to merge with other data sources. Such data is commonly used as an auxiliary form of data and provides a vital source of data for resource-constrained organisations such as local government, charities, and small businesses \citep{ons2021cesusbenefits}. Furthermore, for these organisations, we propose that our embeddings could replace or supplement proprietary geo-demographic data products such as ACORN \citep{caciwebsite} and Mosaic \citep{experianwebsite}, which provide convenient demographic consumer segments at small geographic units.

In order to produce the LiDAR tile embeddings, we use a multi-step unsupervised pipeline. Firstly, using a Convolutional Neural Network model, we extract a set of raw latent features. We subsequently look to reduce the size of the embeddings and segment the embedding space using a combination of Principal Component Analysis (PCA) and K-means clustering. The result is a set of embeddings that are both predictive and interpretable.

We consider how certain factors influence the predictive performance of the embeddings. These can broadly be split into two types. Firstly, we investigate those modelling choices made in the production of the embeddings. Most notably, we review two types of unsupervised embedding methodology approaches, including a `direct' transfer learning approach, which we extend using SimCLR \citep{chen2020simple,chen2020big}: a training framework which uses a form of self-supervised learning. Secondly, for the resulting embeddings, we consider the modelling decisions an analyst end-user might make to optimise the predictive performance. These decisions include the choice of predictive model (i.e.\@ linear or non-linear), embedding size, and how the tile embeddings can be aggregated to larger geographic areas.

To assess our embeddings, we consider the performance in an example supervised/predictive task while also qualitatively assessing the coherence of the clusters/segments. Firstly, for the predictive task, we assess performance by predicting seven domains of deprivation in the Greater London area over small geographical units called Lower Super Output Areas (LSOAs). The seven indices include 1) Income, 2) Employment, 3) Education, 4) Health, 5) Crime, 6) Barriers to Housing \& Services and 7) Living Environment. We chose to benchmark performance using these indices as they are publicly available and cover a wide range of socio-demographic outcomes. This variety allows the embeddings to be assessed in a broad setting since they are intended to be task-agnostic. Importantly, we consider the performance of the embeddings both alone and combined with standard structured data, to assess their potential as an auxiliary source of data. Secondly, we review the clusters/segments produced by our embedding pipeline in terms of the imagery, geo-location and deprivation indices. Via this process, meaning can be injected into the otherwise latent features.

Hence, the aims of our research are as follows:
\begin{enumerate}
  \item To understand if tile embeddings can be used to predict a range of socio-demographic outcomes.
  \item To assess what embedding generation method, parameters and modelling choices lead to more predictive embeddings.
  \item To identify factors relevant to the embedding end-use, including compatibility with non-linear and linear machine learning models, the methods by which the embeddings can be aggregated to larger demographic units, and the impact of embedding size.
  \item To demonstrate how the embedding features can be interpreted by assessing similarities in their visual appearance, demographics and geo-locational attributes.
\end{enumerate}

The results of our work can be used in a variety of machine learning and analytics problems, reducing the time and resources required to integrate remote sensing data. 

The rest of this paper is structured as follows. In section \ref{related_work}, we outline related work and the contributions of this work in relation to the existing literature. Then, in section \ref{methods}, we outline our methods for generating and assessing the embeddings, before describing our data (section \ref{sec:data}) and outlining the experiments (section \ref{experiments}). In section \ref{results}, we review the results, starting with a predictive performance comparison (\ref{results:predictive}), and next demonstrating how the embedding segments can be interpreted using a clustering analysis (\ref{results:clustering}). We then summarise our findings from both a technical and practical perspective in section \ref{conclusion} and highlight limitations of our research and areas of potential further research in section \ref{further_research}.

\section{Related work and contributions}\label{related_work}

This section is structured as follows. Firstly, in section \ref{related_work:applications_with_lidar}, we provide an overview of LiDAR sensing and its typical use cases. Then, we review how alternative types of remote sensing have started to be used in a socio-demographic setting and how this has been made possible, in part, with the introduction of Deep Learning (section \ref{related_work:other_remote_sensing}). We subsequently explore the relevance of unsupervised and self-supervised Deep Learning (section \ref{related_work:self_supervised}), specific types of Deep Learning that allows us to produce task-agnostic embeddings. Then, in section \ref{related_work:geo_demographic}, we outline what geo-demographics is, how it is used and why LiDAR might be relevant to the field.

\subsection{Applications with LiDAR data} \label{related_work:applications_with_lidar}

LiDAR is a method by which distances can be measured. Akin to radar, laser light is emitted from a base point with the time delay between the reflected light providing a measure of distance over a given object or area. LiDAR technology can produce high-resolution imagery and can be used to target a wide array of materials. This capability is particularly advantageous in the environmental sciences when using satellite and airborne LiDAR imaging for surveying, as ground level and above ground features (e.g.\@ buildings, cars, trees) are distinguishable. Examples of applications in the environmental domain include tree species mapping and classification \citep{hamraz2019deep,cao2019estimating,mayra2021tree}, mangrove mapping \citep{li2021mapping}, crop monitoring \citep{lin2021quality}, fire damage assessment \citep{garcia2020evaluating}, landslide detection \citep{syzdykbayev2020persistent} and glacier analysis \citep{telling2017analyzing}. Similarly, the technology has been applied to the urban environment, for example, to understand changes in the urban environment as a result of development \citep{zhou2020lidar} or natural disaster \citep{wang2020extraction}. Furthermore, airborne LiDAR is also prevalent in the archaeological sciences for site detection \citep{de2005using,fernandez2015using,albrecht2019learning, balsi2021preliminary}. While these are all examples of aerial-based LiDAR, the technology is flexible and can also be used for land-based applications. For example, LiDAR is considered one of the enabling technologies for autonomous vehicles \citep{wang2017pedestrian,zermas2017fast,gao2018object}. 

While examples of LiDAR applications are numerous in the environmental and archaeological domains, examples in the social sciences are more limited. Several studies, however, have used metrics and classifications derived from LiDAR to predict or explain socio-economic phenomena. For example, in the field of urban planning, \citet{grove2014ecology} use LiDAR combined with other data sources to produce an Existing Vegetation Cover (EVC) index for private land in New York City. They explore statistical relationships with various socio-economic measures and classifications to enhance targeted land management practices. They demonstrate that vegetation patterns reflect environmental buying behaviours and, among others, lifestyle and life-stage factors. They also show reasonable correlations between the EVC measure and outcomes including; Population Density, Education and Income. These are relevant to our research as they share some overlap with the indices of deprivation we use to assess our embeddings. Similar research has demonstrated that LiDAR-based vegetation indices, including tree cover and remnant vegetation, vary across the socio-economic spectrum, with higher tree coverage present in more advantaged areas of Brisbane, Australia \citep{shanahan2014socio}. Alternative approaches have used LiDAR imaging to derive metrics and classifications that describe building attributes at either a household or area-based level, including volume, footprint, and typologies. Such measures have been shown to be predictive of a range of outcomes including house prices (Syracuse, USA) \citep{lu2013remote}, population density (Denver, USA) \citep{lu2011volumetric}, and socio-demographic indices (Belmopan, Belize) \citep{warth2020prediction}.

While these approaches demonstrate predictive capacity, the major drawback is that they reduce highly complex urban scenes into a single metric or set of metrics. Of course, the advantage to this is that these are simple to interpret, which is particularly important when explicitly reviewing a specific aspect of the urban environment. The downside is that a substantial amount of information is also discarded in the raw data which might be suitable for a given predictive task. However, in many settings that utilise machine learning, sacrificing some interpretability to improve the predictive result is often acceptable. Therefore, our research seeks to address this gap in the literature by distilling complex LiDAR imagery into a set of task-agnostic embeddings using unsupervised Deep Learning approaches, which we subsequently assess against seven domains of deprivation.

\subsection{Socio-demographic applications with other remote sensing data} \label{related_work:other_remote_sensing}

While, for socio-economic analysis, the use of LiDAR combined with Deep Learning approaches remains unexplored, a growing field of research has looked to exploit these techniques with other forms of remote sensing and related imagery. This trend has been primarily driven by the capacity for Deep Learning to both downscale and fuse this complex data \citep{yuan2020deep}. 

Deep Learning approaches can be broadly split into two types: supervised and unsupervised learning. In supervised learning, we would like to predict a labelled outcome ($y$) using a known set of features ($X$), facilitated by a statistical algorithm. This is usually motivated by a desire to predict an unobserved outcome ($y^*$) given a new observable set of input features ($x^*$) \citep{jordan2015machine}. A supervised approach is used by \citet{law2019take} in order to predict house prices in London using a combination of street-view and satellite imagery grid tiles as input. In order to do so, the authors utilise transfer learning, the process by which a model trained on a related task can be `fine-tuned' on the target domain using labelled data, in this instance, with the well-known VGG architecture \citep{simonyan2014very}. Interestingly, they demonstrate that the image inputs can improve predictive accuracy even when combined with standard structured inputs (structural, neighbourhood, and locational housing features) --- which we also seek to demonstrate in our results. Similarly, \citet{suel2021multimodal} combine both street-view and satellite imagery to predict socio-demographic outcomes in London, including income, overcrowding and living environment (quality). They show that the combination of the inputs is better than using the inputs individually. However, unlike \citet{law2019take}, they do not review the predictive performance of the models when used in combination with standard features. \citet{zou2021abandoned} also use a pre-trained VGG as a base model which they fine-tune to classify to classify abandoned houses in the US using street-view imagery alone, achieving 85\% in overall accuracy. Such examples of supervised applications are not restricted to satellite and street-view imagery. \citet{zhao2018tweets} assess the use of nighttime light and social media data (tweet activity) as a means to predict personal income, electric power consumption, and fossil fuel (carbon dioxide) emissions by reviewing their statistical relationships. They do this for relatively large geographic units in the USA: at a state and county level. The authors find both sources of data to be predictive, with the best performing data source being dependent on the target variable and geographic unit.

These studies demonstrate the considerable potential that remote sensing data has in a socio-economic context. However, one challenge with the supervised end-to-end training approaches is that they require granular labelled data to a suitably high resolution. This is not problematic for house price prediction, where both large databases of labelled data are available, and the connection between visual aspects such as neighbourhood attractiveness and house price make end-to-end training viable. There are, however, many applications where this is not the case and where pre-trained embeddings may be helpful, particularly where there is limited labelled data and the link between visual features and the outcome are less obvious.

Unsupervised Deep Learning approaches present a means to extract meaningful feature representations without the presence of labelled data. Furthermore, in some instances, unsupervised techniques can be extended to make use of limited labelled data in what is referred to as semi-supervised learning. For these reasons, recent research in this area mainly focuses on the developing world where fine-grained labels are not available or incomplete. Like for supervised approaches, transfer learning is also commonly adopted in the unsupervised domain. However, for transfer learning, it is desirable that the pre-trained model is contextually similar to the target domain. This prerequisite is particularly challenging in remote sensing as most pre-trained models, regardless of architecture, are trained on the well-known Imagenet database \citep{deng2009imagenet}. While Imagenet is extensive, it differs from aerial type imaging, tending to contain object collections such as animals, cars and buildings. However, that is not to say that transfer learning cannot be utilised even with a significant domain shift, as high-level abstractions can still be relevant. This is demonstrated by \citet{block2017unsupervised}, who utilise a direct transfer learning approach to detect slums using satellite imagery tiles. Their paper proposes a framework for tile segmentation by first extracting features from a ResNet-152 CNN (pre-trained on Imagenet) before performing K-means clustering and dimensionality reduction to segment the images. While also not requiring any labels, the authors show how their approach can lead to interpretable image clusters, which, as they demonstrate for the city of Mumbai, can be predictive of several socio-economic outcomes. \citet{wurm2019semantic} use an alternative approach to segment satellite imagery in order to identify slums. They also apply direct transfer learning, however they do so with a model pre-trained on related imagery. They find transfer learning to significantly improve performance, albeit when transfer task images are of a lower resolution. This is significant as it suggests that models must not just adapt to the domain context, but also to the image resolution.

Similarly to the previous studies, \citet{jean2016combining} used satellite imagery in order to predict poverty levels for five nations located in Africa. However, they extend the direct transfer learning approach by fine-tuning a CNN to the satellite imagery domain with a proxy task, using what is commonly referred to as semi-supervised learning. With limited actual labelled data, the authors first train a model to predict nightlight (night imagery). The model is then subsequently fine-tuned to predict poverty-related outcomes with the available labelled data. \citet{persello2020towards} also use a semi-supervised approach to predict a single socio-economic index reflecting multiple levels of deprivation for neighbourhoods in Bangalore, India. However, rather than using nightlight imagery prediction as a pre-text task, they instead train a model to predict known slum designations.

To conclude, supervised learning with remote sensing data has shown promise for predicting various socio-demographic related outcomes; however, it requires end-to-end model training and available labelled data. In contrast, previous unsupervised and semi-supervised learning applications have shown that aerial imaging can still be used to infer socio-demographic outcomes even with limited or no labelled data. To date, these approaches have either included direct transfer learning or the use of proxy labels/images. However, an alternative approach to learning without labels has recently produced excellent results but is yet to be explored in this context --- self-supervised learning.

\subsection{Advancements in unsupervised representation learning} \label{related_work:self_supervised}

Prior to recent advances in Deep Learning, sophisticated rule-based methods were the dominant means to extract representations from images. These approaches allowed features to be derived in a semi-automated manner, rather than requiring that features be handcrafted entirely. In turn, these representations could be used for a range of downstream tasks such as object detection, object classification, edge detection and image segmentation. Of the rule-based approaches, perhaps the most prevalent method is Scale Invariant Feature Transform \citep[SIFT;][]{lowe2004distinctive} or other methods derived from it such as PCA-SIFT \citep{ke2004pca} and GLOH \citep{mikolajczyk2005performance}. The SIFT approach looks to identify key points within an image, to which a local set of descriptors is then derived for each of these key points. As these descriptors are numerous, and the counts can differ between individual images, there are several extensions to the SIFT methodology that look to condense this information into a smaller number of dimensions, including BOV \citep{sivic2003video}, Fisher Vectors \citep{perronnin2010improving} and VLAD \citep{jegou2010aggregating}. While these classical feature extraction approaches are powerful and computationally efficient, they are underpinned by a rigid set of assumptions and require some degree of modeller involvement with regards to parameter selection which may depend on the application domain. In contrast, Deep Learning approaches are more flexible and capable of learning complex representations directly from the data, without specific domain knowledge or modeller involvement. Furthermore, Deep Learning approaches have been shown to outperform classical approaches across a range of supervised and unsupervised tasks \citep{luus2015multiview,tian2017l2,lin2018unsupervised,ranccon2019comparison}.

Approaches that use Deep Learning for unsupervised representation learning can be broadly classified as either generative or discriminative. While generative approaches attempt to model the input space or generate new instances, discriminative approaches instead learn the boundaries between classes. In an unsupervised context, however, labelled data is not available, so discriminative approaches look to generate pseudo labels, typically using features of the input images. Discriminative approaches have arguably proved to be more popular as they are more intuitive and less computationally expensive than generative approaches, which tend to model the data at a pixel level. 

Earlier discriminative approaches typically involved a pre-defined image manipulation with a Deep Learning model tasked to detect a given random transformation, for example, image reconstruction \citep{doersch2015unsupervised, noroozi2016unsupervised}, colourisation \citep{zhang2016colorful} and rotations \citep{gidaris2018unsupervised}. Such an approach has been applied with LiDAR data. Specifically, \citet{wang2019pixel} recently extracted pixel-wise features for land coverage classification, albeit with a Random Forest model rather than a Deep Learning model.

More recently, a type of discriminative learning, called contrastive learning, has emerged that encourages similar images to yield similar latent representations. At the time of writing, the state of the art in contrastive learning is SimCLR \citep{chen2020simple,chen2020big}, which rather than being a specific type of model architecture, is proposed as a framework. The framework operates by creating a series of duplicate images and applying various image transformations to them. These paired images are then fed into a training process that is designed to produce similar representations for them. The framework has yielded impressive results, with the latest iteration even outperforming training with labels (supervised) under some conditions \citep{chen2020big}.

Despite the promise of self-supervised learning approaches, this new generation of unsupervised learning is relatively underexplored in remote sensing, where it is arguably highly suited given the abundance of unlabelled imaging data. We address this gap by applying the SimCLR framework as part of a pipeline to derive meaningful embeddings and segments.

\subsection{The relevance of geo-demographic data} \label{related_work:geo_demographic}

Geo-demographics is defined as ``the use of detailed categorical and continuous data on small areas defined and selected according to categorisations that are set up to discern high/low incomes, levels of education and deprivation, status indicators such as taste and purchasing patterns, and other data'' \citep{geodemographics}. Although typically, this type of data is associated with targeting and segmenting customers for marketing purposes \citep{mitchell1994role}, its potential applications are vast, serving both commercial and public interest. Examples include allocating public policing \citep{ashby2005geocomputation}, public health intelligence \citep{abbas2009geodemographics,petersen2011geodemographics}, public transport planning \citep{liu2020understanding}, financial service branch management \citep{birkin1998gis} and deciding restaurant locations \citep{muller1994geodemographics}.

Perhaps one of the most important data sources underpinning geo-demographic studies in the UK is the national census survey. As well as having national coverage, the data produced by this survey contains a wide array of rich socio-demographic information and is open source. Due to the complexity of working with this information in raw form, however, several commercial offerings simplify the data by generating convenient customer segments. Perhaps the most well-known geo-demographic products in the UK are `ACORN' and `Mosaic' offered by CACI and Experian, respectively. These products utilise national census data, enriching it with proprietary and other open data sources and offering segmentations at a more fine-grained level (household or postal units) than that available from the census publication (output areas). Open source segments referred to as Output Area Classifications are also available for the 2001 \citep{vickers2007creating} and 2011 \citep{gale2016creating} census, published by the UK Office for National Statistics (ONS). However, these purely utilise census data and are published at larger geographical units than the commercial offerings.

Alternatively, LiDAR elevation tiles form a rich source of open data (in the UK) and are produced at a high geographic resolution. Furthermore, the data is updated and published annually. Despite this, elevation tiles have not been explored in either a commercial or open-source setting, which we look to address with our tile embeddings and resulting segments.

\section{Method}\label{methods}

Our research's primary aims are to generate task-agnostic tile embeddings and understand whether those are predictive of a range of outcomes and remain so when used in conjunction with structured variables. Hence, the methods we apply can broadly be split into two categories: `embedding generation' and `embedding use'. The former is concerned with the modelling decisions made to produce the embeddings, whereas the latter relates to the end-user's modelling decisions to optimise the use of the embeddings. The following two subsections outline the approach applied for each of these two elements; \ref{methods:emb_gen} (embedding generation) and \ref{methods:emb_use} (embedding end use).

\subsection{Embedding generation}\label{methods:emb_gen}

\begin{figure}[!htb]
\centering
\scalebox{0.65}{
  \includegraphics[width=0.99\linewidth]{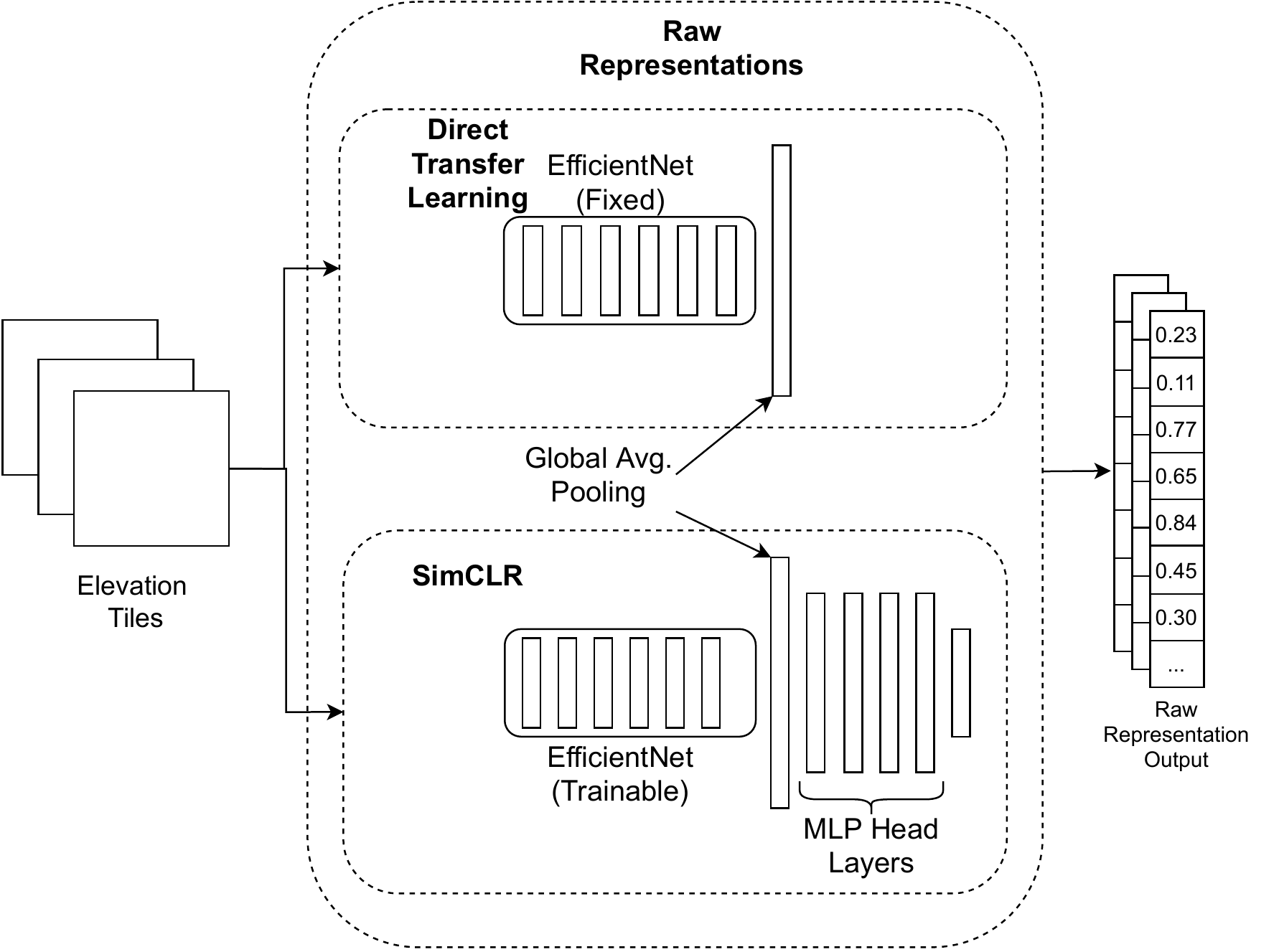}}
  \caption{Pipeline: Raw representations. This part of the pipeline is concerned with extracting a flat 1-dimensional feature representation per tile via the Direct Transfer Learning approach or the SimCLR training framework. Note that while the Direct Transfer Learning model has a single output per tile, for SimCLR, we consider four hidden layers on the MLP head.}
  \label{fig:RawRepresentations}
\end{figure}

The first stage in the generation of the embeddings is to extract representations from a Deep Learning model which we refer to as the `base model'. We do so in an unsupervised manner using two approaches outlined in this section: 1) direct transfer learning, which we extend using 2) the SimCLR training framework (see figure \ref{fig:RawRepresentations}).

\subsubsection{Direct transfer learning}
Transfer learning is a process by which the weights of a model trained on a related task are re-used in a different application. Typically the base model undergoes a process of `fine-tuning' to the intended task, usually in a supervised context (with labels). This approach has been widely adopted both in the image processing and natural language processing fields. 

We use an EfficientNet implementation as our base model, reflecting the current state of the art in supervised Deep Learning image processing \citep{tan2019efficientnet}. The EfficientNet was published with seven variations increasing in size and parameters (B0 to B7). We implement the smallest version (B0), pre-trained on same-sized images using the Imagenet dataset. A global pooling of the final layers of the base model is used to produce a raw feature vector of size 1280 per tile. As the task is unsupervised, there is no further fine-tuning of the base model to the LiDAR tiles; hence we refer to this as a `direct' transfer learning approach. This is broadly aligned with the method of extraction adopted by \citet{block2017unsupervised}.

\subsubsection{SimCLR training framework}
\begin{figure}[!htb]
\centering
\scalebox{1}{
  \includegraphics[width=0.98\linewidth]{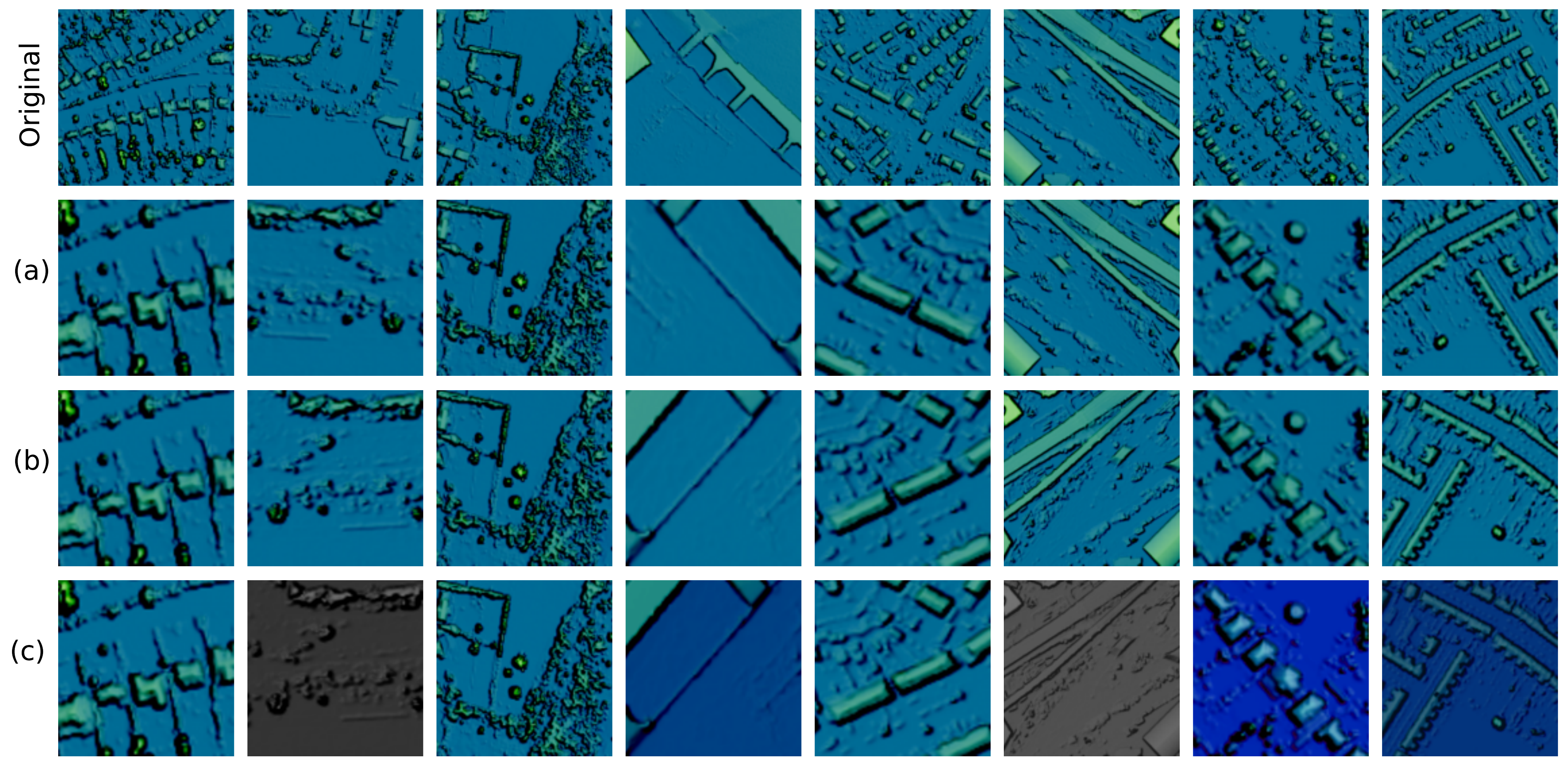}}
  \caption{Example of augmented LiDAR tiles which form input into the SimCLR model training. The original LiDAR image tiles are on the top row while the layered stochastic augmentations are presented in subsequent rows. The random transformations include: (a) zooming and cropping, (b) image flipping, (c) colour distortion and blurring.}
  \label{fig:AugmentedTiles}
\end{figure}

The second approach we take to extract raw representations is to undertake additional training of the base model using the LiDAR tiles. This additional fine-tuning should, in theory, outperform a direct approach as the base model can adapt to the LiDAR image tiles, which differ from the types of images the pre-trained model has been exposed to. In a supervised context, the base model would be fine-tuned using labels for a specific task --- estimating the seven domains of deprivation in our instance. However, a supervised approach would be counter-intuitive for our research goals as we do not intend to focus on a single downstream task but create embeddings that are task-agnostic and can thus be used for a wide range of (other) supervised learning problems. We, therefore, need to train the model using an unsupervised learning approach, to which we apply the SimCLR training framework. The framework uses self-supervised learning to create dummy labels, effectively converting the task to a supervised one. This is done by randomly sampling images from the dataset and generating transformed image copies using random augmentations (Figure \ref{fig:AugmentedTiles}), e.g.\@ cropping, rotations, zooming and blurring. Such image augmentations are also commonly applied in supervised learning to prevent over-fitting; however, the motivation differs slightly. With SimCLR, the training objective is to encourage mapping transformations of the same image closely in the embedding space. To do so, the authors use Normalized Temperature-scaled Cross Entropy Loss, or NT-Xent. The purpose of the loss is to attract transformed duplicates while repelling representations of other images within the same training batch. This self-supervised process allows the model to learn latent features that distinguish the image inputs, producing generic features that can be used for the downstream task. The described approach is what the SimCLR authors refer to as `unsupervised pretraining'; they then further demonstrate how this pretraining can be extended using transfer learning and semi-supervised learning using limited labelled data. For our analysis, however, we only consider the unsupervised pretraining elements of the framework, as it is this part that allows us to produce the task-agnostic embeddings for LiDAR data.

The base model's architecture remains the same as that of the direct approach (EfficientNet B0), to which we append a Multilayer Perceptron (MLP) network consisting of four dense layers of size 512 and a projection head. The final dense layer (the projection head) produces the representation output for the loss function, while each of the preceding dense layers of the MLP will be considered for the embedding output. A size of 512 was initially selected as it provides a manageable upper bound for the final embedding. We also validate this choice in the predictive results where we assess the impact of embedding size on performance.  Our configuration is selected based on the findings from the second iteration of the SimCLR paper \citep{chen2020big}. The authors found that a deeper MLP head can significantly improve the outcome, especially when the base model has fewer parameters --- as the EfficientNet does. However, the authors also note that the raw representations' performance can vary according to which layer abstraction is used, with earlier layers tending to capture higher-level features that are more useful for the downstream task. We, therefore, report the result of all four dense layers preceding the projection head of the SimCLR model.

\subsubsection{Embedding post-processing}

\begin{figure}[!htb]
\centering
\scalebox{0.75}{
  \includegraphics[width=0.99\linewidth]{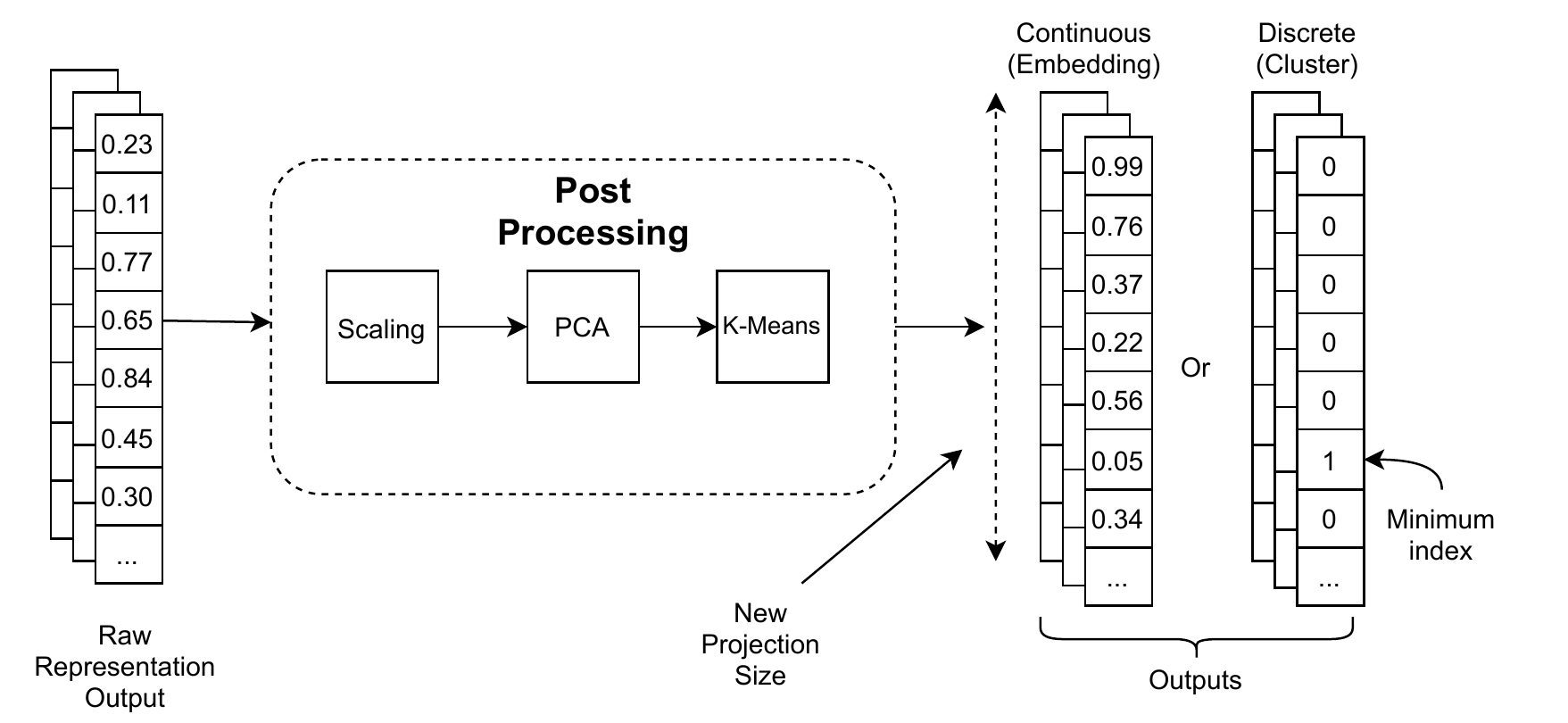}}
  \caption{Pipeline: Post-processing. The second part of the embedding processing pipeline takes the raw feature representation produced by the Deep Learning models. This additional step maps the raw embedding into the target dimension size by setting the \emph{K} parameter of K-Means. While the continuous output space (cluster distances) can be used as feature embeddings for a machine learning model, the space can also be easily discretised by taking the index of the minimum distance for feature interpretation.}
  \label{fig:PostProcessing}
\end{figure}

The raw representations go through three further processing stages to arrive at the final embeddings: Scaling, Principal Component Analysis (PCA) and K-means (see Figure \ref{fig:PostProcessing}). These additional steps form a standard machine learning approach that allow us to simultaneously explore the impact of different embedding sizes while also outputting the representations in a more interpretable form.

Firstly, the vector representations (raw embeddings) are standardised, which is a requirement for the later steps. Second, PCA is a linear form of Singular-Value-Decomposition that projects the standardised embeddings into a lower-dimensional space to reduce noise and redundancy in the embeddings. The \emph{n\_components} (number of components) parameter is tuned for each embedding so that the minimum number of components is obtained whilst explaining at least 99\% of the variance. PCA could form the final dimensionality-reduction stage in the process; however, we opt to use K-means to reach the target embedding size. K-means is a popular clustering algorithm that seeks to partition the input data (here, PCA transformed tile representations) into $K$ discrete groups, where $K$ is a predetermined number of clusters. By setting $K$ to the target embedding size, the transformed output for each tile embedding is a vector with the distance from each of the $K$ cluster centres. Smaller distances from a centroid suggest the tile is more representative of the cluster. The benefit of this approach is that the output embedding values are interpretable as by identifying typical features of each cluster, we can add meaning to the latent embedding. Furthermore, the embeddings can be easily discretised by taking the embedding index with the minimum distance and allocating the tile to the corresponding cluster. We wish to understand the trade-off between the size of the embeddings and predictive performance, and therefore trial embeddings sized 4, 8, 16, 32, 64, 128, 256 and 512.

\subsection{Embedding end use}\label{methods:emb_use}

\begin{figure}[!htb]
\centering
\scalebox{0.65}{
  \includegraphics[width=0.99\linewidth]{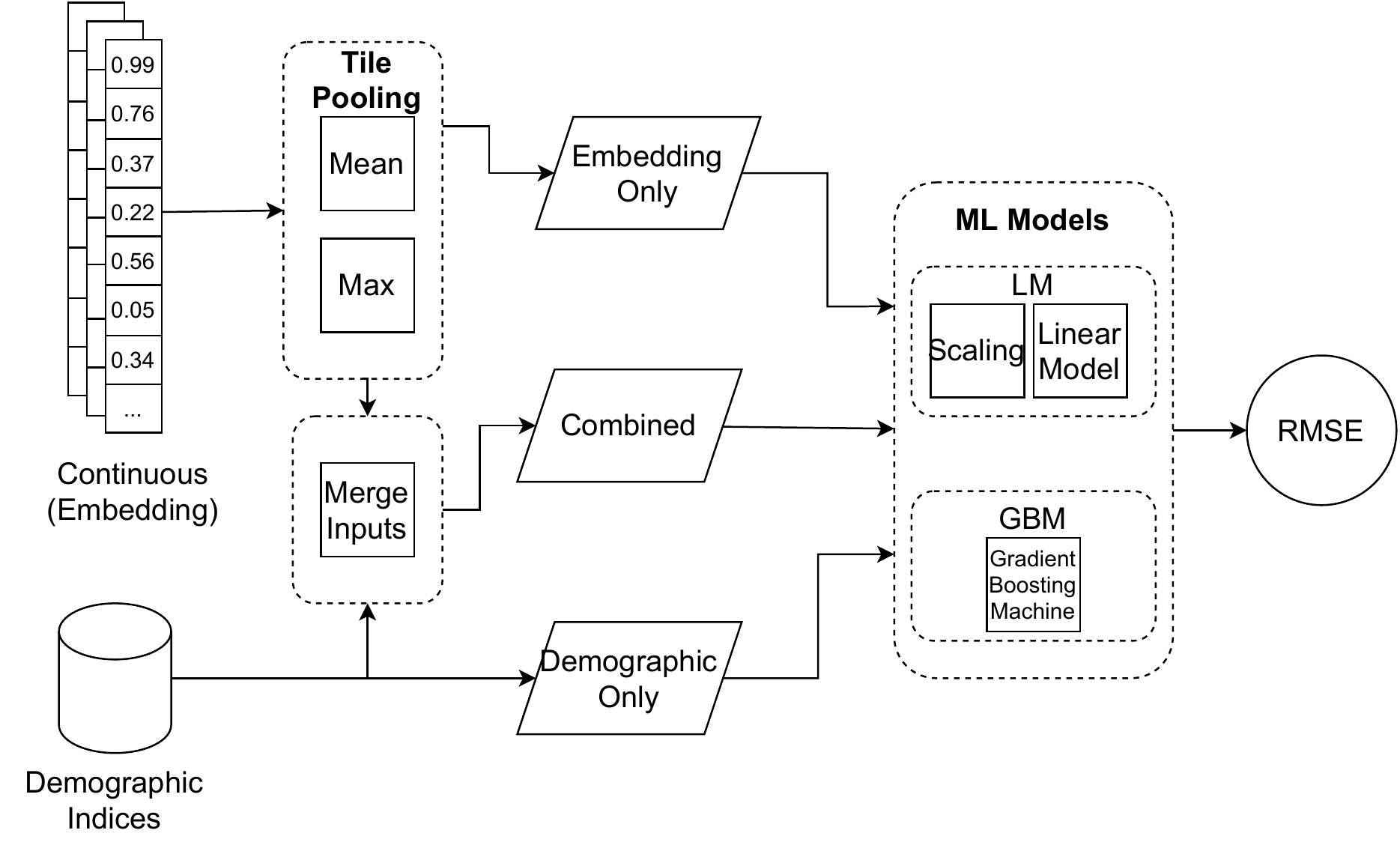}}
  \caption{Pipeline: End use. The final part of the pipeline moves from embedding generation to embedding use. This figure shows how the two data inputs (Indices and Embeddings) feed into the machine learning models to predict the target index. Furthermore, it shows how, for the embeddings, there is an additional processing step that looks to aggregate the tile level embeddings to an LSOA level.}
  \label{fig:MLpipe}
\end{figure}

This subsection describes how the derived embeddings can be used to predict the indices of deprivation (see Figure \ref{fig:MLpipe}). As well as testing the value of the task-agnostic embeddings, this also demonstrates how the embeddings might be used in a typical machine learning pipeline.

\subsubsection{Aggregating embeddings}
It is a requirement for the predictive task that there is a single representation per LSOA, the geographic area at which the indices of deprivation are published. Given that an LSOA contains multiple tiles, the embeddings must be allocated to an LSOA and aggregated.

The tile embeddings are assigned to the LSOAs according to the location of the tile centroid, ensuring that the embeddings are allocated only once. While this approach is reasonably rigid, it provides a robust means to assess the predictive capacity of the embeddings as information cannot be leaked between training, validation and test sets. In practice, an end-user may consider relaxing this constraint and use alternative methods such as area-based weighting, or, dictionary-based methods which we will describe later.

While embedding aggregation is a requirement for our purposes, it is also practically relevant for other potential use cases when using the embeddings to represent larger geographic units. We, therefore, trial two types of aggregation, using \emph{mean-pooling} and \emph{max-pooling} approaches. Applied to the K-means embedding representations, these approaches are intuitively equivalent to taking the mean and maximum Euclidean distance from each \emph{K} cluster centre. Consider a simplified example, with just two clusters, one representing blocks of flats while the other consists of green open space. Mean-pooling would effectively ask, `what is the average representation of flat blocks and green space over all tiles in a given area?', while max-pooling would ask `for all tiles in a given area, what was the maximum representation of green space and flat blocks regardless of how many tiles they occur in?'.

The \emph{mean-pooling} and \emph{max-pooling} methods present an accessible means to aggregate the embeddings for a wide range of use cases. Alternative dictionary based approaches from the classic computer vision literature which utilise hard cluster assignment may also be applicable in some cases, including BOV, Fisher Vectors and VLAD. One issue with these approaches is, however, that they can be sensitive to the interplay between the embedding size (i.e.\@ the number of clusters) and the number of tiles within a given area. For example, choosing too many clusters combined with too few tiles per area could result in a very sparse representation and a significant loss of information. Moreover, these methods introduce additional implementation complexity in that they require a greater degree of expert knowledge. For these reasons, we opt to limit the scope to only these pooling methods. 

\subsubsection{Machine learning models}
We wish to understand how the embeddings perform in a linear and non-linear setting for the predictive task. This is of practical interest as the more flexible non-linear models can often provide better performance; however, this usually comes at the cost of increased training time, tuning complexity and lesser transparency when compared to linear models. We, therefore, apply two models, a Linear least-squares Model (LM) and a non-linear Gradient Boosting Machine (GBM). These models reflect two ends of the bias-variance tradeoff and are typical examples of machine learning models that an end-user might adopt. While there are many non-linear algorithms, we have selected the GBM as it has been shown to perform well in many settings. For the linear model, we use a Lasso implementation that applies L1 penalisation and is efficient in our setting where there are a potentially large number of features. For the GBM, we use the LightGBM implementation of the algorithm \citep{ke2017lightgbm}. 

For the LM model, we apply five fold cross-validation on the training data to trial 100 variations of the alpha regularisation parameter. While the LM model is relatively fast and straightforward to train, the GBM is more complex with increased training times and more hyper-parameters. Therefore, we use the Optuna framework for this model to select an appropriate set of hyperparameters \citep{optuna_2019}. Specifically, we implement the Tree-structured Parzen Estimator (TPE) algorithm \citep{bergstra2011algorithms}, a Bayesian optimiser that iteratively looks at sampling the search space to minimise the validation set root mean squared error (RMSE) with a budget of twenty trials. Furthermore, to control over-fitting for the GBM model, we also implement early stopping which prevents the model growing once performance stops improving in the validation set.

\subsection{Data}\label{sec:data}
\subsubsection{Deprivation Indices}
We assess our embeddings using the English Indices of Deprivation dataset published by the UK Office for National Statistics (ONS) \citep{ministry_of_housing_2019}. These are particularly relevant for our assessment as the seven core domains reflect a wide array of social-economic outcomes and will give a good indication as to the applicability of the embeddings. A description of the seven domains is presented in Table \ref{DomainsDescrition}. The indices were published at an LSOA level consisting of 200 to 400 dwellings or 400 to 600 persons. To demonstrate the approach, we chose to collect data from the Greater London area, as this provides a reasonable sized sample with a variety of geographical and socio-economic features. In this area of interest, there are 4400 LSOAs in total.

As the seven domains are not within the same numerical ranges, we re-scale the indices between 0 and 100 using a min/max procedure to be able to cross-compare the RMSE scores for the predictive task across the domains.

\begin{table}[!htb]
    \centering
    \caption{Seven domains of deprivation described by the Office for National Statistics \citep{ministry_of_housing_2019}}
    \begin{small}

    \begin{tabularx}{\linewidth}{lX}
    \toprule
    \textbf{Domain} & \textbf{Description}\\
    \midrule
    1) Income   & Measures the proportion of the  population experiencing deprivation relating to low income\\
    \midrule
    2) Employment   & Measures the proportion of the working age population in an area involuntarily excluded from the labour market\\
    \midrule
    3) Education   & Measures  the lack of  attainment and skills in the local population\\
    \midrule
    4) Health   & Measures the risk of premature death and the impairment of quality of life through poor physical or mental health\\
    \midrule
    5) Crime   & Measures the risk of personal and material victimisation at local level\\
    \midrule
    6) Barriers to Housing   & Measures the physical and financial accessibility of housing and local services\\
    \midrule
    7) Living Environment   & Measures the quality of both the `indoor' and `outdoor' local  environment\\
    \bottomrule
    \end{tabularx}
    \end{small}
    \label{DomainsDescrition}
\end{table}

\subsubsection{LiDAR data}
The LiDAR elevation data are acquired from the EDINA Digimap data portal \citep{edina} and are also available from the GOV website under an Open Government Licence. The data is available at a 0.25cm resolution; however, we opt for the 1m resolution as it provides better coverage whilst still providing a high enough resolution to identify detailed features such as cars, trees, fences and urban furnishings. The raw data is available as a Digital Surface Map (DSM), which provides a measure of the ground level, and a Digital Terrain Map (DTM), inclusive of above-ground features. The DSM raster tiles are subtracted from the DTM tiles to get the final elevations and provide a normalised 3D map. While the DTM data can be used directly, this additional step ensures that geo-locational clues are not explicit within the embeddings; i.e.\@ the embeddings should be independent of the absolute ground level. We then subset the raster tiles into series of 200x200m discrete non-overlapping grids, the same size as applied by \citet{block2017unsupervised}, producing a total of 44000 grid tiles for the extended Greater London area.

\subsection{Experiments}\label{experiments}

For the predictive task, we wish to understand if the embeddings are predictive on their own and if they are predictive when used together with structured variables. We therefore consider three subsets of the data: 1) the demographic data only ($D_{demographic}$) which is formed of the remaining six (of seven) indices, 2) the elevations embeddings only ($D_{embeddings}$) and 3) the combined data ($D_{combined}$) which uses both $D_{demographic}$ and $D_{embeddings}$ as input. In order to assess the predictive performance, we use the Root Mean Squared Error (RMSE) metric, with a reduction in the RMSE between the $D_{demographic}$ and $D_{combined}$ results indicating an improvement with the introduction of $D_{embeddings}$. 

The validation and test data consist of LSOAs made up of two selection strategies similar to that applied by \citet{law2019take}, using an area-based hold-out and a random hold-out. Firstly, for the area-based samples, we set aside an entire Local Authority District (LAD), a larger geographic area consisting of multiple LSOAs. Secondly, for the random hold-out, we use LSOAs randomly sampled across the Greater London area. Conceptually, the area-based hold-out is more challenging as the model will not have been exposed to neighbouring and presumably similar LSOAs. However, this sample is potentially limited in terms of the overall variation of elevation tiles. Conversely, the random hold-out subset will have greater coverage; but the model will have the advantage of seeing embeddings from closer proximity. By combining the two approaches, we can draw on the pros of both. 

\begin{figure}[!htb]
\centering
\scalebox{0.85}{
  \includegraphics[width=0.99\linewidth]{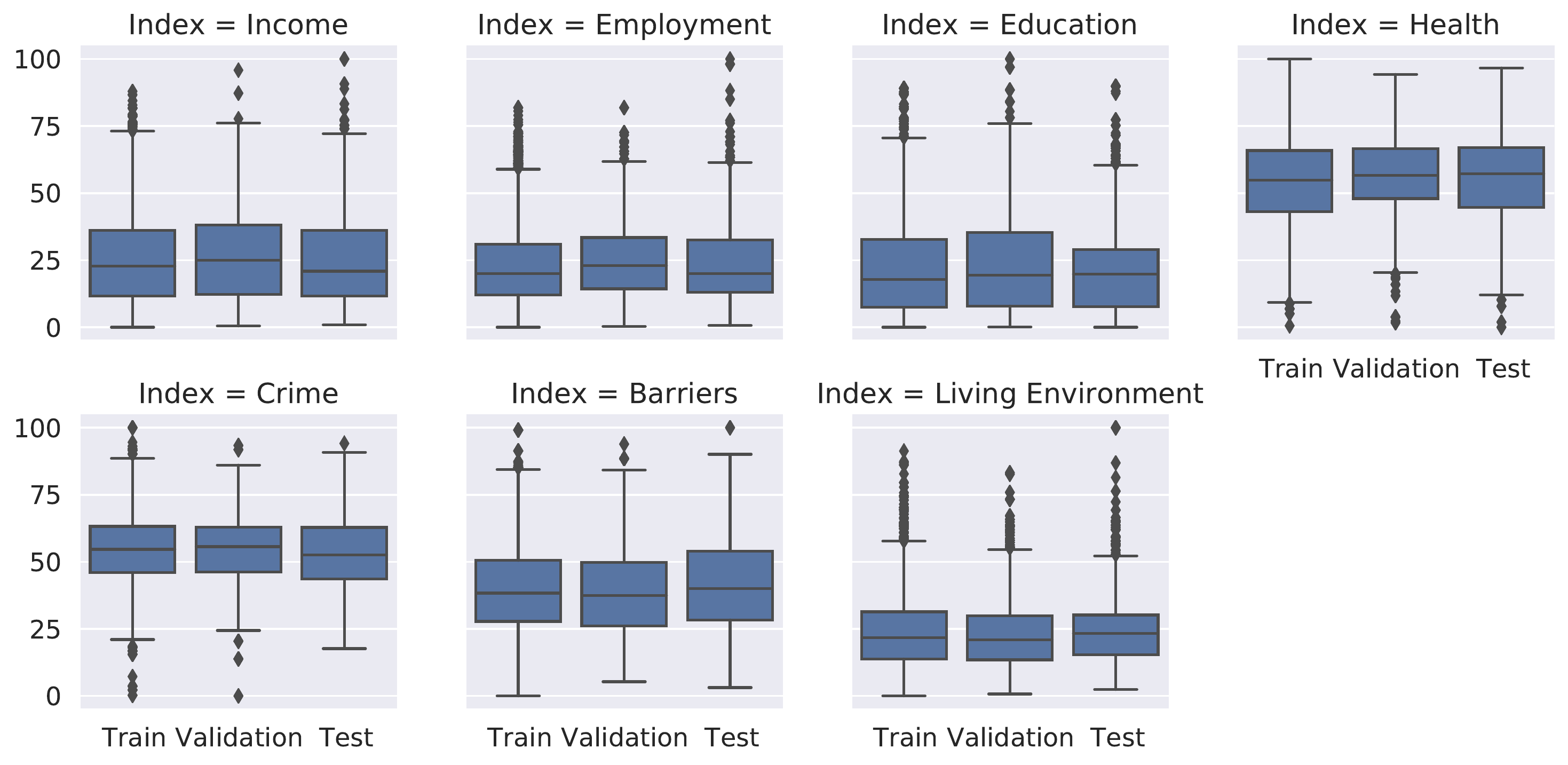}}
  \caption{Deprivation indices distribution}
  \label{fig:areaCharachter}
\end{figure}

To train the machine learning models, we use a 60\%/20\%/20\% Train/Validation/Test split of the LSOAs. As explained, the validation and test sets consist of a single LAD area, with the remaining proportion of the amalgamated set made up of the randomly sampled LSOAs. For example, the selected LAD area contains enough LSOAs to make up 46\% of the required test set (20\% of all LSOAs). The remaining 54\% then consists of the randomly sampled LSOAs. Figure \ref{fig:areaCharachter} shows the final distribution of the deprivation indices across the Training, Validation and Test sets, from which we can conclude they are sufficiently similar to validate performance.

\section{Results}\label{results}
\subsection{Predictive task results}\label{results:predictive}
The core performance results for the predictive task are presented in Table \ref{CoreResults}. 
\begin{table}[!htb]
\centering
\caption{Core RMSE results for the predictive task on the test set. The best performing model for each domain (i.e.\@ with the lowest RMSE) is in \textbf{bold} format. \% Improvement shows the percentage increase in the performance between a model built using $D_{demographic}$ and one using $D_{combined}$, for the corresponding embedding and model type. We select our largest embedding size (512) with mean-pooling for these core results, and for the SimCLR embeddings, we use the first layer (L1).} 
\vspace{+0.5mm}
\scalebox{0.77}{
\begin{tabular}{lcccccccc}
\toprule
\multicolumn{2}{c}{\textbf{Data Subset}}                                          & \textbf{$D_{demographic}$} & \multicolumn{2}{c}{\textbf{$D_{embedding}$}}                                            & \multicolumn{2}{c}{\textbf{$D_{combined}$}}                                             & \multicolumn{2}{c}{\textbf{\% Improvement}}                                          \\ \midrule
\multicolumn{2}{c}{\textbf{Embeddings}}                                           & \textbf{-}              & \textbf{\begin{tabular}[c]{@{}c@{}}Direct\\ Transfer\end{tabular}} & \textbf{SimCLR} & \textbf{\begin{tabular}[c]{@{}c@{}}Direct\\ Transfer\end{tabular}} & \textbf{SimCLR} & \textbf{\begin{tabular}[c]{@{}c@{}}Direct\\ Transfer\end{tabular}} & \textbf{SimCLR} \\ \midrule
\textbf{Domain}                              & \multicolumn{1}{l}{\textbf{Model}} & \textbf{}               & \textbf{}                                                          & \textbf{}       & \textbf{}                                                          & \textbf{}       & \textbf{}                                                          & \textbf{}       \\ \midrule
\multirow{2}{*}{\textbf{barriers}}           & \textbf{GBM}                       & 11.9                    & 13.2                                                               & 13.1            & 11.4                                                               & \textbf{10.8}  & 4\%                                                                & 9\%             \\
                                             & \textbf{LM}                        & 12.6                    & 13.5                                                               & 12.9            & 12.3                                                               & 12.0            & 2\%                                                                & 4\%             \\ \midrule
\multirow{2}{*}{\textbf{crime}}              & \textbf{GBM}                       & 10.1                    & 11.2                                                               & 10.9            & 9.7                                                                & 9.6             & 4\%                                                                & 5\%             \\
                                             & \textbf{LM}                        & 10.1                    & 11.2                                                               & 10.7            & 9.6                                                                & \textbf{9.5}   & 4\%                                                                & 6\%             \\ \midrule
\multirow{2}{*}{\textbf{education}}          & \textbf{GBM}                       & 10.5                    & 12.2                                                               & 12.6            & 9.1                                                                & \textbf{8.9}   & 14\%                                                               & 16\%            \\
                                             & \textbf{LM}                        & 11.1                    & 12.3                                                               & 12.0            & 9.5                                                                & 9.1             & 14\%                                                               & 18\%            \\ \midrule
\multirow{2}{*}{\textbf{employment}}         & \textbf{GBM}                       & 4.8                     & 12.4                                                               & 12.7            & 4.8                                                                & 4.7             & 0\%                                                                & 2\%             \\
                                             & \textbf{LM}                        & 4.7                     & 12.2                                                               & 12.1            & \textbf{4.5}                                                      & \textbf{4.5}   & 4\%                                                                & 5\%             \\ \midrule
\multirow{2}{*}{\textbf{health}}             & \textbf{GBM}                       & 8.2                     & 11.5                                                               & 12.6            & 7.6                                                                & \textbf{7.5}   & 7\%                                                                & 9\%             \\
                                             & \textbf{LM}                        & 9.3                     & 11.5                                                               & 11.6            & 8.5                                                                & 8.7             & 8\%                                                                & 6\%             \\ \midrule
\multirow{2}{*}{\textbf{income}}             & \textbf{GBM}                       & 5.0                     & 13.2                                                               & 13.7            & 4.6                                                                & 4.6             & 8\%                                                                & 8\%             \\
                                             & \textbf{LM}                        & 5.1                     & 13.2                                                               & 12.9            & 4.6                                                                & \textbf{4.5}   & 10\%                                                               & 10\%            \\ \midrule
\multirow{2}{*}{\textbf{living environment}} & \textbf{GBM}                       & 10.8                    & 9.7                                                                & 9.8             & 9.3                                                                & 9.1             & 13\%                                                               & 16\%            \\
                                             & \textbf{LM}                        & 11.2                    & 9.7                                                                & 9.5             & 9.1                                                                & \textbf{8.9}   & 19\%                                                               & 21\%            \\ \bottomrule
\end{tabular}}
\label{CoreResults}
\end{table}

We see that, for all deprivation domains, the best performing models were those inclusive of the embedding, with observed improvements of up to 21\% compared to using the $D_{demographic}$ data alone. While both embedding types lead to an RMSE improvement in the combined model, the SimCLR embeddings notably outperform the direct transfer approach in six out of seven domains, the sole exception being the employment domain where RMSE performance was the same (4.5). The largest improvement of 21\% was observed for the living environment domain. This is somewhat intuitive given that LiDAR is likely to directly capture many relevant features here, such as measures of urban density or the availability of open green space. Furthermore, in this domain, we see that even the embeddings alone ($D_{embedding}$) could produce better performance than using the $D_{demographic}$ subset as input. A considerable increase of 18\% is also observed for the education domain using the SimCLR embeddings, reflecting that urban environment features are highly predictive of local education levels. Interestingly, these results are broadly aligned with those of \citet{grove2014ecology}, who found that vegetation indices derived from LiDAR data exhibited the strongest correlations with Income and Education in New York City.

As for the performance differences between the two machine learning models, GBM is the best performing method in the barriers, education and health domains, whereas the LM model gives superior results on the crime, employment income and living environment domains. However, in all instances, the differences between the two are marginal, meaning the embeddings are already extracting most of the knowledge, providing sufficient information for a less sophisticated linear model to incorporate in its final prediction. This is a significant finding: we can use complex non-linear embeddings to do the ``heavy lifting'' while the simpler linear models can be used for the prediction. This was precisely our goal, as the embeddings would facilitate deployment by smaller organisations without the technical capacity to implement complex models.

\begin{figure}[!htb]
\centering
\scalebox{1}{
  \includegraphics[width=0.99\linewidth]{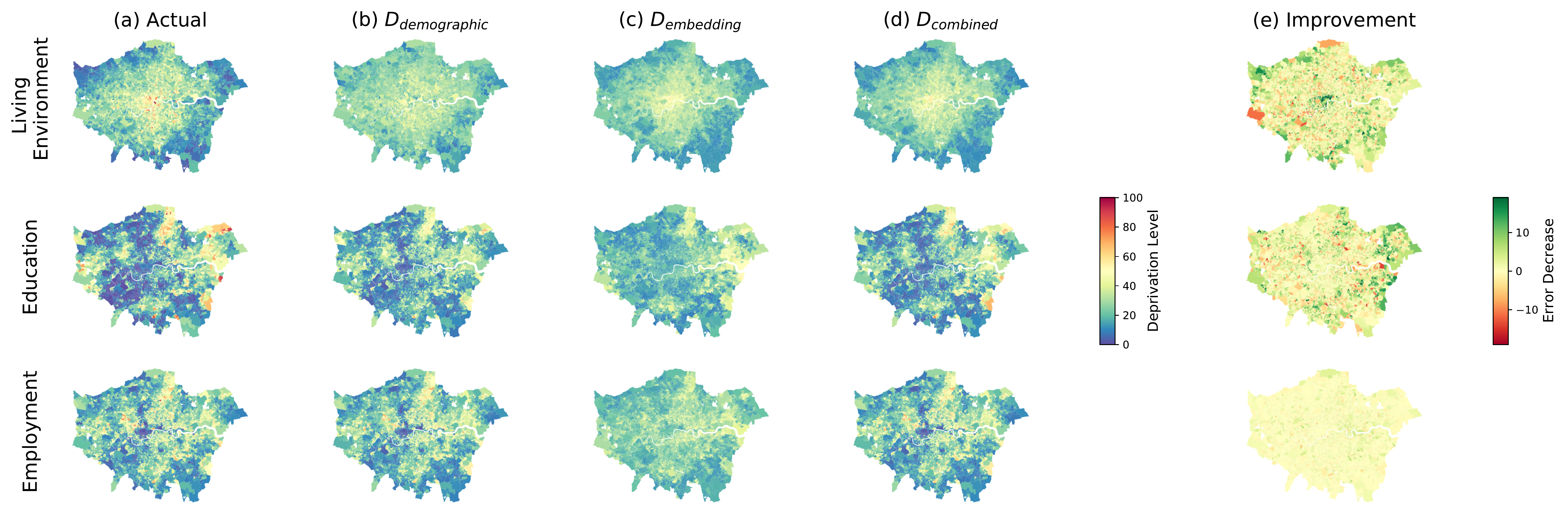}}
  \caption{LSOA level predictions. This figure shows the actual deprivation levels for three indices (a) against the predicted values, for the three subsets $D_{demographic}$ (a), $D_{embedding}$ (b) and $D_{combined}$ (c). This results set uses the GBM model, and for the $D_{embedding}$ and $D_{combined}$ results, the SimCLR embeddings are adopted using the largest embedding size (512) with mean-pooling. The improvement plots (e) show the geographical distribution of the error improvement in the $D_{combined}$ subset compared to the $D_{demographic}$ subset.}
  \label{fig:GeoError}
\end{figure}

Figure \ref{fig:GeoError} shows the distribution of the predicted values (b, c, d) against the actual levels of deprivation (a) for three of the indices, as well as the error improvement (e) when when comparing the model using $D_{combined}$ against using only $D_{demographic}$. These three indices were selected as they show a range in the $D_{combined}$ improvements, including two examples where the embeddings produce notable performance increases (Living Environment and Education) as well an example with lesser improvement overall (Employment). This plot is reproduced for all seven indices in \ref{app:error}. Interestingly, we see that the $D_{embedding}$ predictions (b) are capable of broadly capturing the geographical distribution of the actuals (a); however, they may not fully capture the magnitude of the deprivation levels until combined with the demographic data (d). 

When reviewing the error decreases (e) for the Living Environment and Education domains, we see that the improvements are geographically dispersed and are not concentrated in urban, suburban or rural areas. Furthermore, the areas of improvement also vary according to the target domain. These two findings are significant as they suggest that the embeddings are adaptable across both a range of target outputs and terrains. Additionally, for the Employment domain where performance improvements were relatively lower in the $D_{combined}$ subset (2\%-5\%), the moderate error decreases are also evenly distributed over the area. This observation is also promising in a predictive setting, suggesting that even when the embeddings are less helpful, the predictions are stable and do not inject unwanted noise.

\begin{figure}[!htb]
\centering
\scalebox{0.98}{
  \includegraphics[width=1\linewidth]{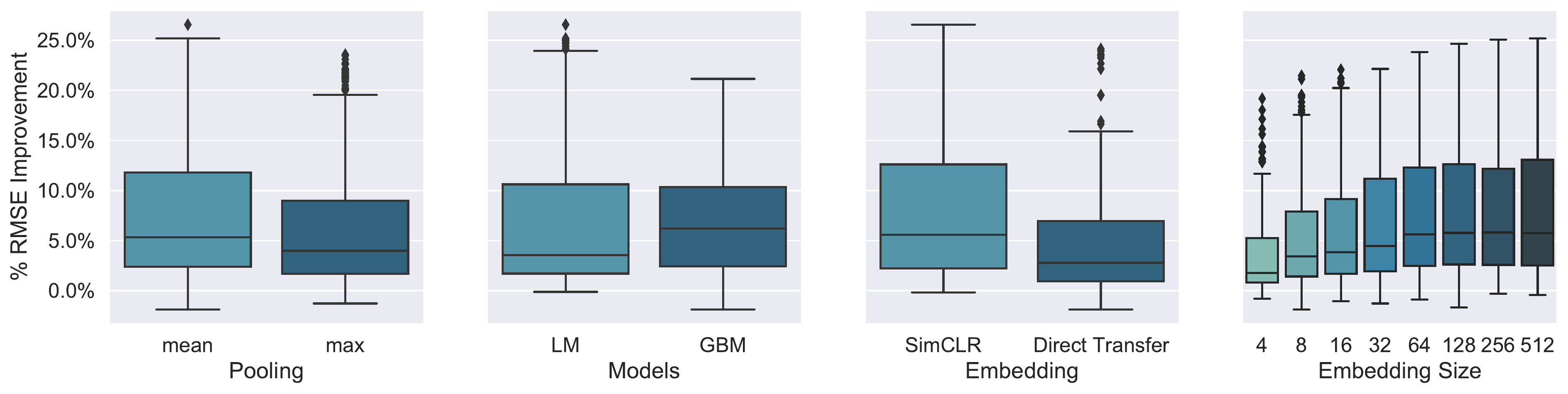}}
  \caption{RMSE improvement by model configuration}
  \medskip
  \tiny
  \label{fig:ModelConfigPlot}
\end{figure}

The next set of results helps us further understand how performance varies with changes in the modelling configuration. Figure \ref{fig:ModelConfigPlot} shows the aggregate RMSE percentage improvement, grouped according to the pooling procedure, model choice, embedding type and embedding size.

When aggregating the tiles to LSOA level, while both mean- and max-pooling procedures lead to an RMSE improvement, mean-pooling notably outperforms max-pooling for our embeddings (see the leftmost plot in Figure \ref{fig:ModelConfigPlot}). The pooling method is an important consideration for an end-user working with areas larger than the tile level --- as is the case for the domains of deprivation. This difference makes sense as mean-pooling represents all tiles through the average, while max-pooling only reflects the most extreme case. In other words, as we are looking to represent the LSOA, mean-pooling produces a more meaningful representation of this wider area compared to max-pooling.

With regards to model selection, improvements are observed for both the non-linear GBM and the LM. While GBM often benefits more from including the embeddings (showing a higher median improvement), there are some configurations where the simpler LM proves the more appropriate choice. 

In terms of the choice of embeddings, as also seen in Table \ref{CoreResults}, the additional fine-tuning using SimCLR training provides clear performance benefits over the Direct Transfer approach. The superior performance of the SimCLR embeddings is somewhat expected due to the significant domain shift from the Imagenet images that the base model (EfficientNet) was trained on to the LiDAR images. In such settings, SimCLR's self-supervised learning strategy is designed precisely to adapt the base model to the new target domain.

When reviewing the predictive performance across the embedding sizes (i.e.\@ $n\_{clusters}$), surprisingly, even the inclusion of just four features --- our smallest embedding size --- leads to an improvement in all experiments. While increasing the embedding size leads to greater improvements in the RMSE, further increases beyond size 64 produce fast diminishing returns (rightmost plot in Figure \ref{fig:ModelConfigPlot}). 

\begin{figure}[!htb]
\centering
\scalebox{0.8}{
  \includegraphics[width=1\linewidth]{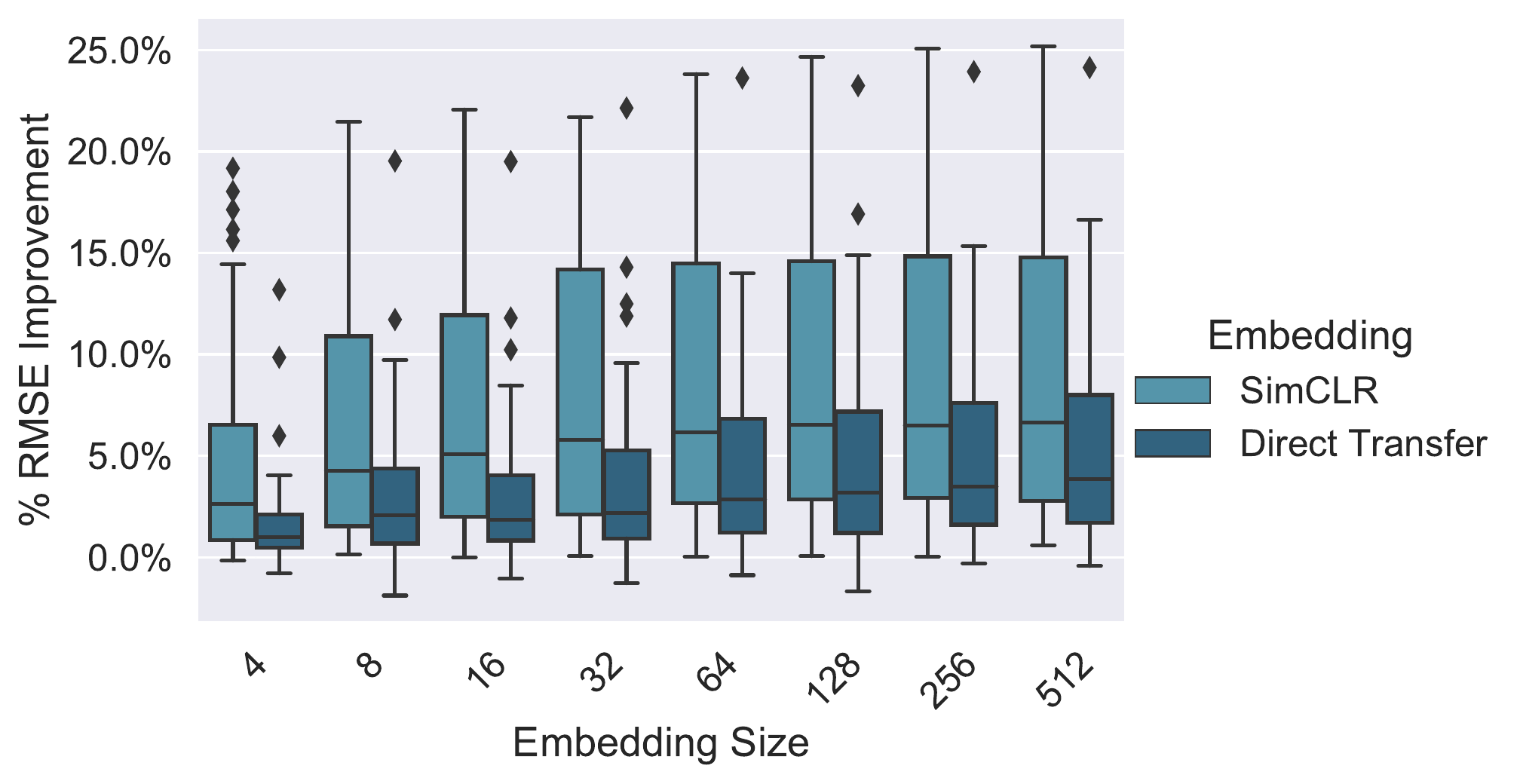}}
  \caption{RMSE improvement by model configuration}
  \medskip

  \label{fig:ModelConfigPlot_Combination}
\end{figure}

While Figure \ref{fig:ModelConfigPlot} demonstrates the impact on the observed RMSE improvement for the individual elements, Figure \ref{fig:ModelConfigPlot_Combination} further explores the interaction between two key components: the embedding choice and embedding size. As shown earlier, the SimCLR model produces larger improvements overall. However, there is some evidence as well that the application of the SimCLR embeddings also reduces the required embedding size. This can be observed in Figure \ref{fig:ModelConfigPlot_Combination} where, for the Direct Transfer approach, there is an almost continuous improvement seen when increasing the embedding size up to 512, whereas for the SimCLR embeddings, there is little further RMSE improvement from size 64 onward. This might suggest that the additional fine-tuning using the SimCLR framework leads to a more efficient representation of the LiDAR tiles. This is intuitive as the base model was trained on Imagenet with a vast array of different images and would thus need a larger capacity to capture this variability. Therefore, it may be that, through a process of self-supervision, not only new information is acquired about the target domain (LiDAR), but redundant information is also discarded.

\begin{figure}[!htb]
\centering
\scalebox{0.6}{
  \includegraphics[width=1\linewidth]{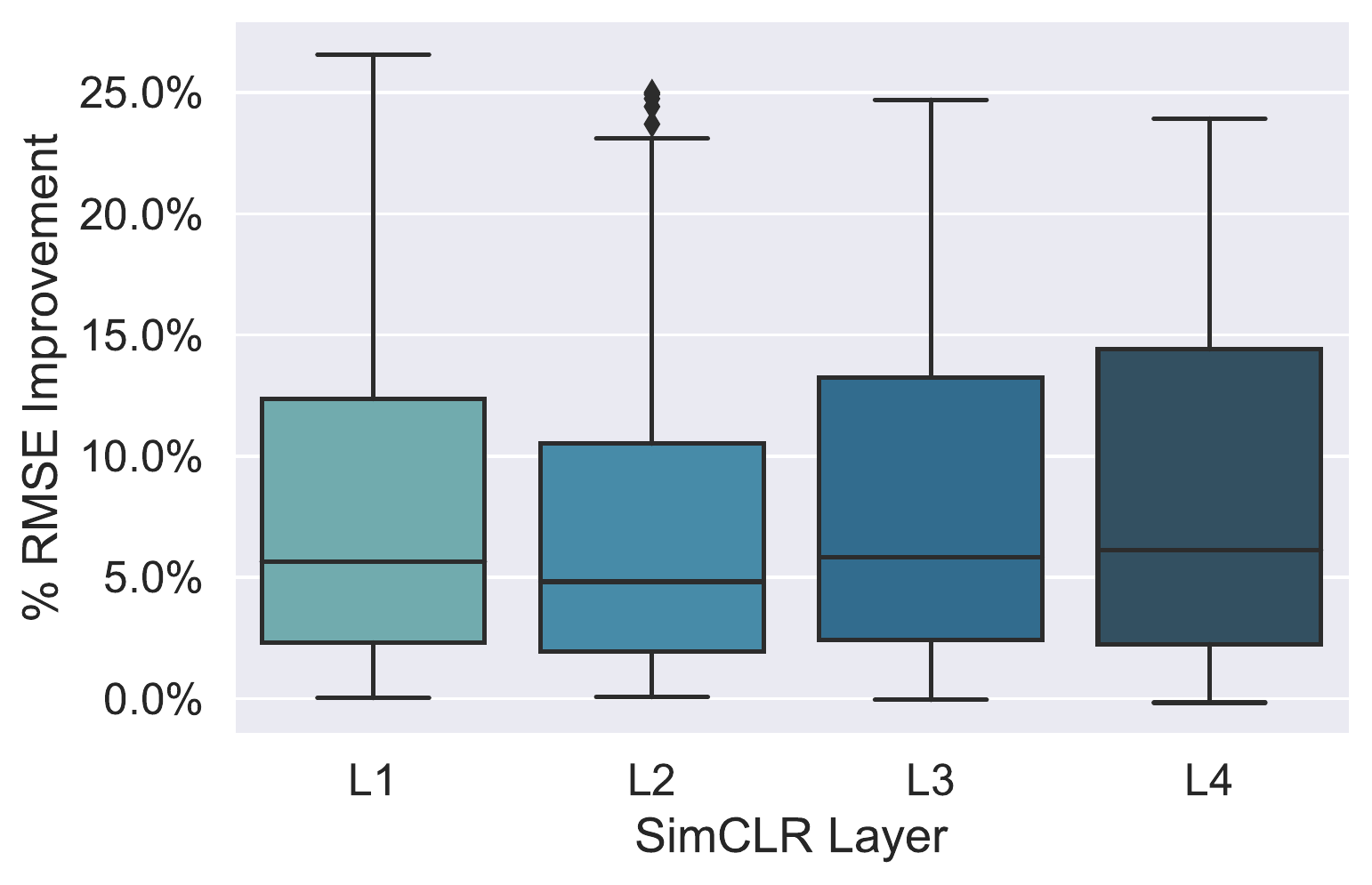}}
  \caption{RMSE improvement by SimCLR Layer}
  \medskip
  \label{fig:SimCLR_Layer_Choice}
\end{figure}

For the SimCLR framework specifically, we also explore which layers of the dense head produce more predictive embeddings. We do this as it has been demonstrated that representation performance can vary according to which layer preceding the final dense layer of the model is selected, with evidence suggesting that earlier layers capturing higher-level abstractions may be more useful for a downstream task \citep{chen2020big}. Our results, however, show that while Layer 1 does result in some of the best performance gains (see the extended tail of the leftmost box plot in Figure \ref{fig:SimCLR_Layer_Choice}), the difference in performance between the layers is small.

\subsection{Clustering task} \label{results:clustering}
In the previous section, we demonstrated that the LiDAR embeddings are predictive of the seven domains of deprivation, even with the socio-demographic features included. This section shows how our embedding pipeline, using a Deep Learning model followed by K-Means clustering, adds value over the raw output embeddings. In addition to reducing the number of dimensions, therefore numerically simplifying the representation, the K-means process also allows the embedding space to be characterised by identifying common features within each cluster and injecting interpretable meaning into the otherwise latent embedding.

While for the predictive task, we considered up to size 512 in the embedding dimension, for this exercise, we use four clusters. This value was selected as it produced the highest Silhouette Score (0.21), thus providing a good trade-off between individual cluster cohesion and separation between the clusters. Also, using relatively few clusters (four) allows the representations to be conveniently described. Note that we only consider the SimCLR embeddings as they produced higher silhouette scores and outperformed the direct transfer embeddings on the predictive task. 
To assign meaning to the resulting clusters, we consider three sources of data: the tile images themselves, the geographic cluster representation and the levels of deprivation. 

\begin{figure}[!htb]
\centering
\scalebox{0.9}{
  \includegraphics[width=1\linewidth]{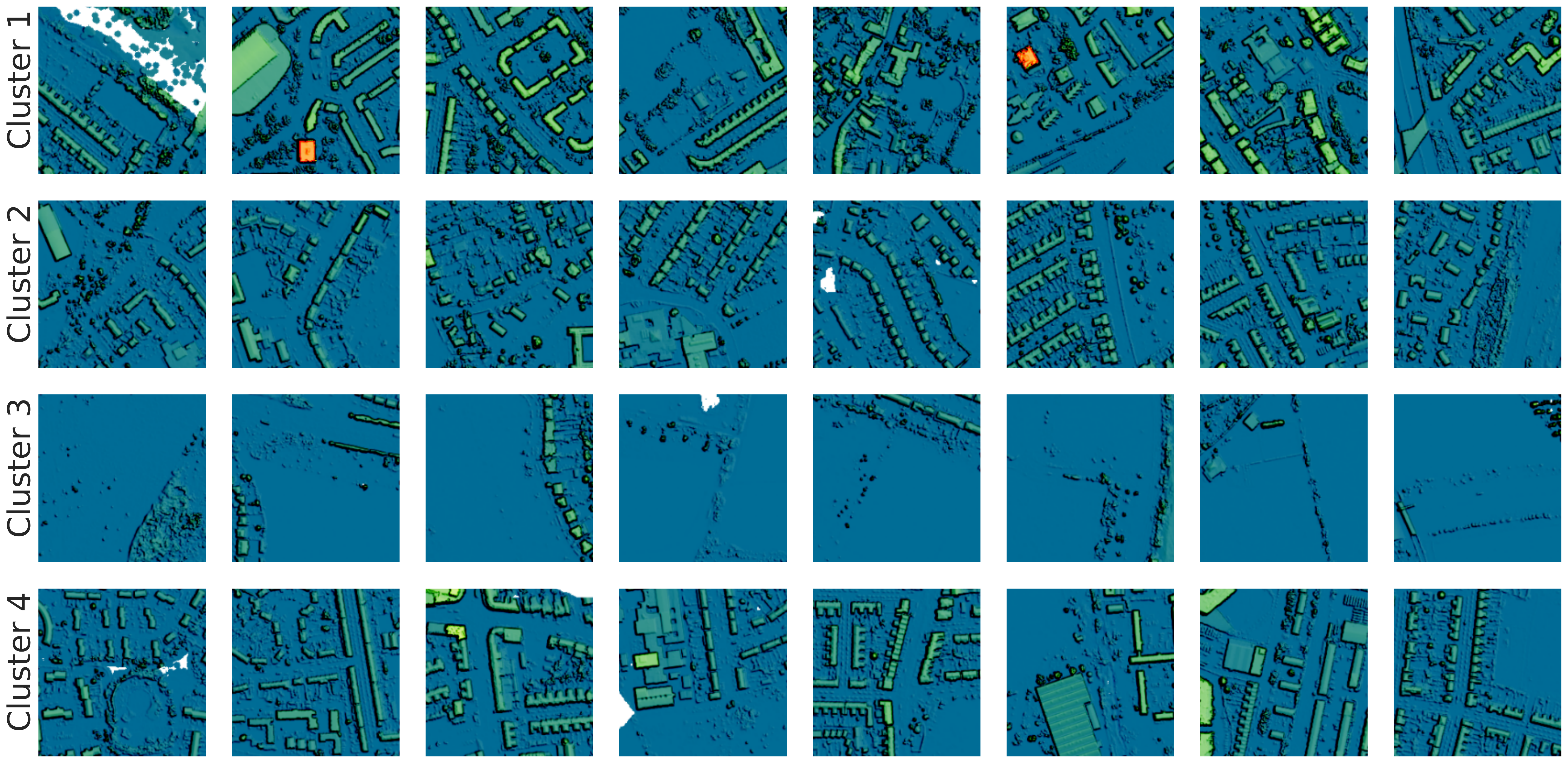}}
  \caption{Representative tile images for four clusters}
  \label{fig:ClusterTiles}
\end{figure}

First, a random sample of representative images from within each cluster is presented in Figure \ref{fig:ClusterTiles}. From these, it becomes evident that there are distinguishable features for each cluster. For example, Cluster 1 is characterised by having a higher density of high-rise buildings indicating higher levels of urbanisation. Conversely, Cluster 3 contains few buildings and is predominantly open space with greenery, including trees and hedges, indicating a more rural setting. Cluster 2 and Cluster 4 are similar in some respects as both seem to be capturing a residential setting. There are, however, some significant differences between the two. While Cluster 2 contains a higher density of detached housing, green features and land boundaries (fencing), Cluster 4 includes flats and larger buildings that might indicate industrial or office use.

\begin{figure}[!htb]
\centering
\scalebox{0.95}{
  \includegraphics[width=1\linewidth]{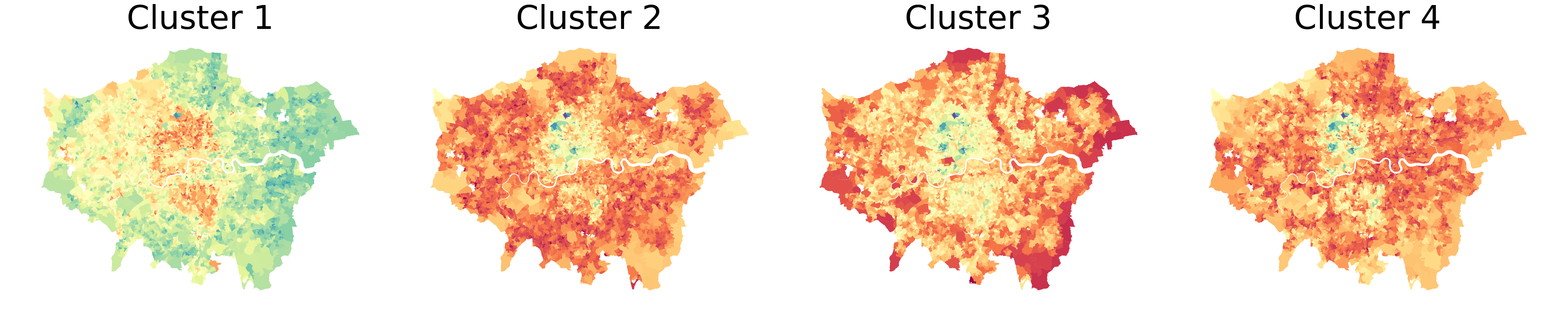}}
  \caption{Cluster representation of LSOAs in Greater London (red/green colour scale denoting high/low mean cluster representation)}
  \label{fig:ClusterMap}
\end{figure}

Secondly, Figure \ref{fig:ClusterMap} shows the mean cluster representation for each Lower Super Output Area (LSOA) on a map of Greater London. The colour red in this plot indicates where the mean euclidean distance from the cluster is smaller and the LSOA is, therefore, more representative of the corresponding cluster. Intuitively, Cluster 1 (urban) is highly concentrated in the city centre, while Cluster 2 (rural) is highly represented in the `green' areas on the city's outskirts. On the other hand, Cluster 2 and Cluster 4 have lower representation in the city centre and tend to be concentrated in the middle belt, suggesting that these are suburban areas.

\begin{figure}[!htb]
\centering
\scalebox{0.8}{
  \includegraphics[width=1\linewidth]{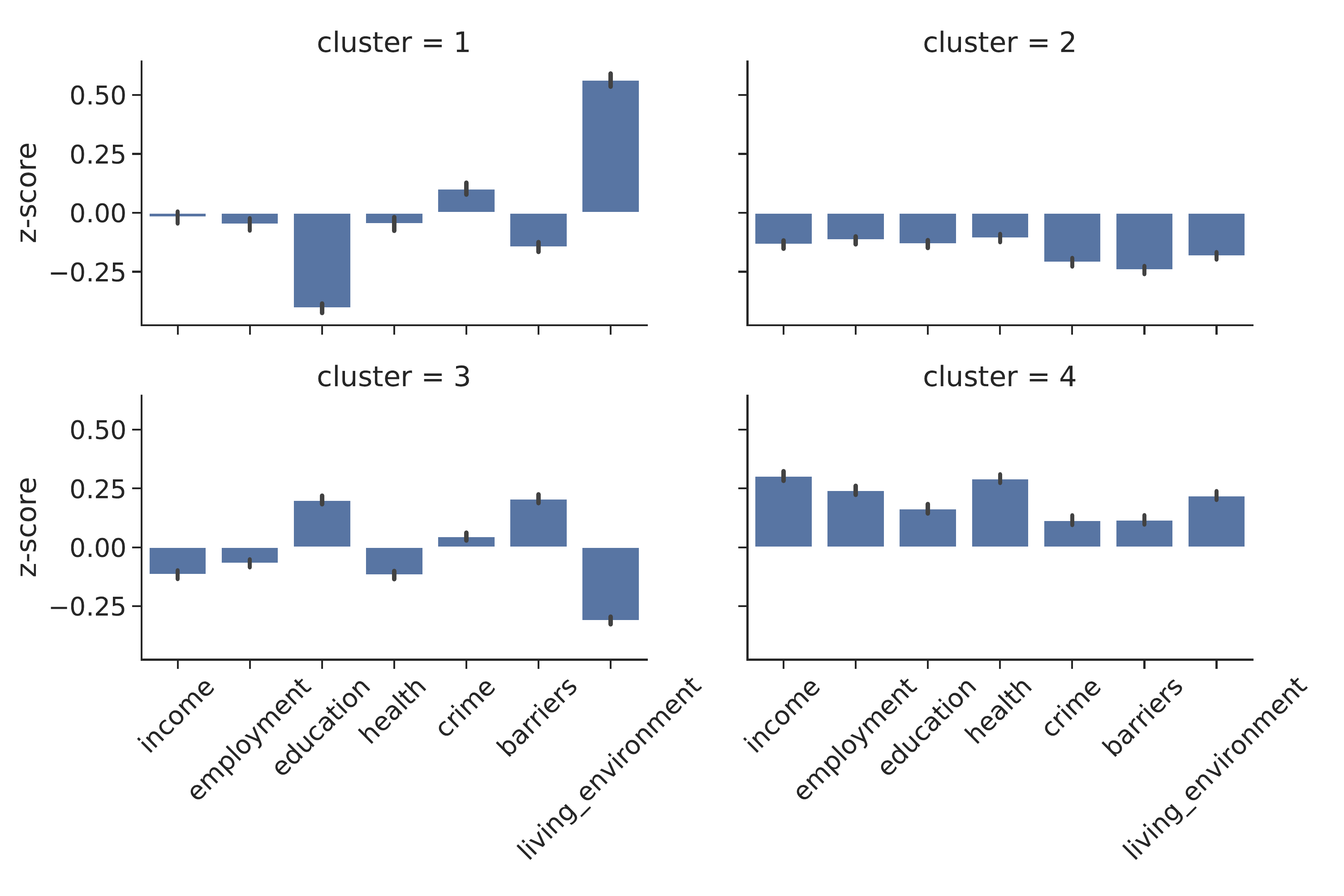}}
  \caption{Normalized average deprivation levels associated with each cluster}
  \label{fig:ClusterBarChart}
\end{figure}

Thirdly, we can also explore why the embeddings are predictive of the seven domains of deprivations, by identifying differences between the four clusters. Figure \ref{fig:ClusterBarChart} thus shows the variation between the deprivation statistics for each cluster, higher values indicating increased levels of deprivation. Note that these indices have first been transformed to z-scores to introduce a more uniform scale for comparison purposes.

As shown by the upper-left plot, Cluster 1 (urban) has very high levels of deprivation in the living environment domain while very low levels in the education domain (i.e.\@ those living there tend to be highly educated). This may go some way to explaining why the predictive performance for our embeddings was highest in these domains, as the urban density appears highly reflective of these two outcomes. This is in part confirmed as Cluster 3 (rural) has an inverse relationship with both. Interestingly, despite Cluster 2 and Cluster 4 sharing similarities in terms of location (suburban) and visual features (residential housing), the two are very different in terms of levels of deprivation. Cluster 2 has below average levels of deprivation across the seven domains, while Cluster 4 has above average levels of deprivation. This is intuitive given that Cluster 2 shared features associated with increased wealth, including more detached housing and private open space marked with property boundaries.

To summarise, we have demonstrated in this section how cluster characterisation can be used as a tool to explore and define the embedding features via our clustering pipeline, even with relatively few clusters. In practice, this holds the potential to further segment the embedding space into increasingly homogeneous groupings for visual interpretation.

\section{Reflection and conclusions}\label{conclusion}
Our research has sought to demonstrate that Deep Learning approaches can be used to derive convenient embeddings from elevation tiles, which can later be seamlessly used for prediction purposes. We have considered the performance of our embeddings in both a supervised and unsupervised context. For the supervised task, we used our embeddings to predict seven domains of deprivation, reflecting a wide array of socio-economic markers. For the unsupervised task, we sought to assess if our embeddings could form coherent clusters that can be conveniently interpreted. Our approach to developing embeddings can be broadly split into two aspects. First, we considered the technical aspects of designing the elevation embeddings to optimise performance, i.e.\@ selection of embedding type and layer choice. Secondly, we reviewed practical aspects concerned with how an end-user might seek to implement the embeddings by selecting aggregation methods, embedding size and machine learning model choice.

\subsection{Technical reflections}
To derive embeddings from the elevation data, we considered two methods of representation abstraction, both using Deep Learning approaches. First, using a direct transfer learning approach, we used the EfficientNet model, pre-trained on the ImageNet task. We then extended this method using the SimCLR training framework, a self-supervised approach to fine-tune EfficientNet to the LiDAR elevation tiles. The SimCLR model consists of the same EfficientNet base, extended with a four-layered non-linear MLP head. Our results demonstrate that while both embedding approaches can yield improved RMSE results across all the domains, when used in conjunction with the other demographic features, the performance is notably improved using the SimCLR approach with performance uplifts of up to 21\%. Furthermore, a review of the geographic dispersion of the error suggests that the embeddings are adaptable to both the terrain and target domain. We would therefore conclude that our embeddings are suitable to be used for a wide array of downstream tasks where socio-demographic type data is useful. We also reviewed the impacts of using different layers from the SimCLR model and find performance is similar across the layers and domains. 

As well as demonstrating predictive performance, we were able to show using four clusters how the embedding space can be interpreted. The clusters were homogeneous in terms of visual appearance and tended to be geographically co-located in either urban, suburban or rural locations. We also reviewed differences between clusters regarding the seven domains of deprivation and found each to have a distinct demographic profile.

\subsection{Practical reflections}
Our tile embeddings are produced for 200x200m grid tiles. However, these tile embeddings had to be aggregated to an LSOA level to assess the predictive performance on the domains of deprivation. While a requirement for our given dataset, this reflects how the tile embeddings might be used in practice when combined with other data collected for larger geographic units. We trialled two aggregation types, mean-pooling and max-pooling, and found that, for our embeddings, mean-pooling outperforms max-pooling in six of the seven domains. This intuitively suggests that how the latent features appear on average in the area is more important than the maximum representation of each feature. We also sought to explore how end-user choices related to embedding size and type of machine learning model influenced the predictive performance, and found some interplay between the two. In our core results set, which used the largest embedding size (512), we found the choice between a linear model and non-linear GBM model arbitrary, with similar performance between the two. This is significant as it suggests that our embeddings can be used with relatively simple models and are therefore suitable for our target users --- smaller organisations with limited capacity to utilise complex models. In practice, however, if modelling complexity is not problematic, then both non-linear and linear models should be trialled as performance is likely to depend on the use case and the types of features used in conjunction with the elevation embeddings. For the choice of embedding size, we find that larger embeddings tend to lead to better RMSE performance, stabilising after size 64 for the SimCLR embedding. However, even with the smallest embedding size (four), we still observe substantial performance increases over using the demographic features alone.

\section{Limitations and further research} \label{further_research}
Our research has demonstrated how task-agnostic embeddings can be derived from elevation data using unsupervised deep learning. Given the success of the SimCLR embeddings applied to the elevation tiles, for further research, we suggest a greater exploration of the approach, for example, by reviewing different types of base models. \citet{chen2020big} found that with ResNet models, having more parameters was helpful for self-supervised learning. We have used a relatively small EfficientNet implementation, partially mitigated using a large projection head; however, larger models may reasonably lead to even better performance. More broadly, further research may look to conduct a wider ablation study which also considers performance relative to pixel-wise deep learning approaches, and, classical image feature extraction approaches such as SIFT with its associated variations. Additionally, other design choices could be explored including the impact of varying the tile sizes and alternative methods of aggregating tiles to larger geographic units.

As a secondary task, we sought to explore how our embeddings could also provide tile clusters/segmentation, similar to \citet{block2017unsupervised}. We did so using K-means, but other clustering approaches may potentially produce better clusters. In addition to this, though out of scope for our research, further interpretation of the embedding features would be of interest for social science research so as to understand why the tiles are predictive of socio-demographic outcomes.

\section*{Acknowledgements}
This work was supported by the Economic and Social Research Council [grant number ES/P000673/1]. The last author acknowledges the support of the Natural Sciences and Engineering Research Council of Canada (NSERC) [Discovery Grant RGPIN-2020-07114]. This research was undertaken, in part, thanks to funding from the Canada Research Chairs program.

The authors acknowledge the use of the IRIDIS High Performance Computing Facility, and associated support services at the University of Southampton, in the completion of this work.

{\small
\bibliography{references}}

\newpage
\appendix
\section{LSOA Level Predictions}\label{app:error}
\begin{figure}[!htb]
\centering
\scalebox{1}{
  \includegraphics[width=0.99\linewidth]{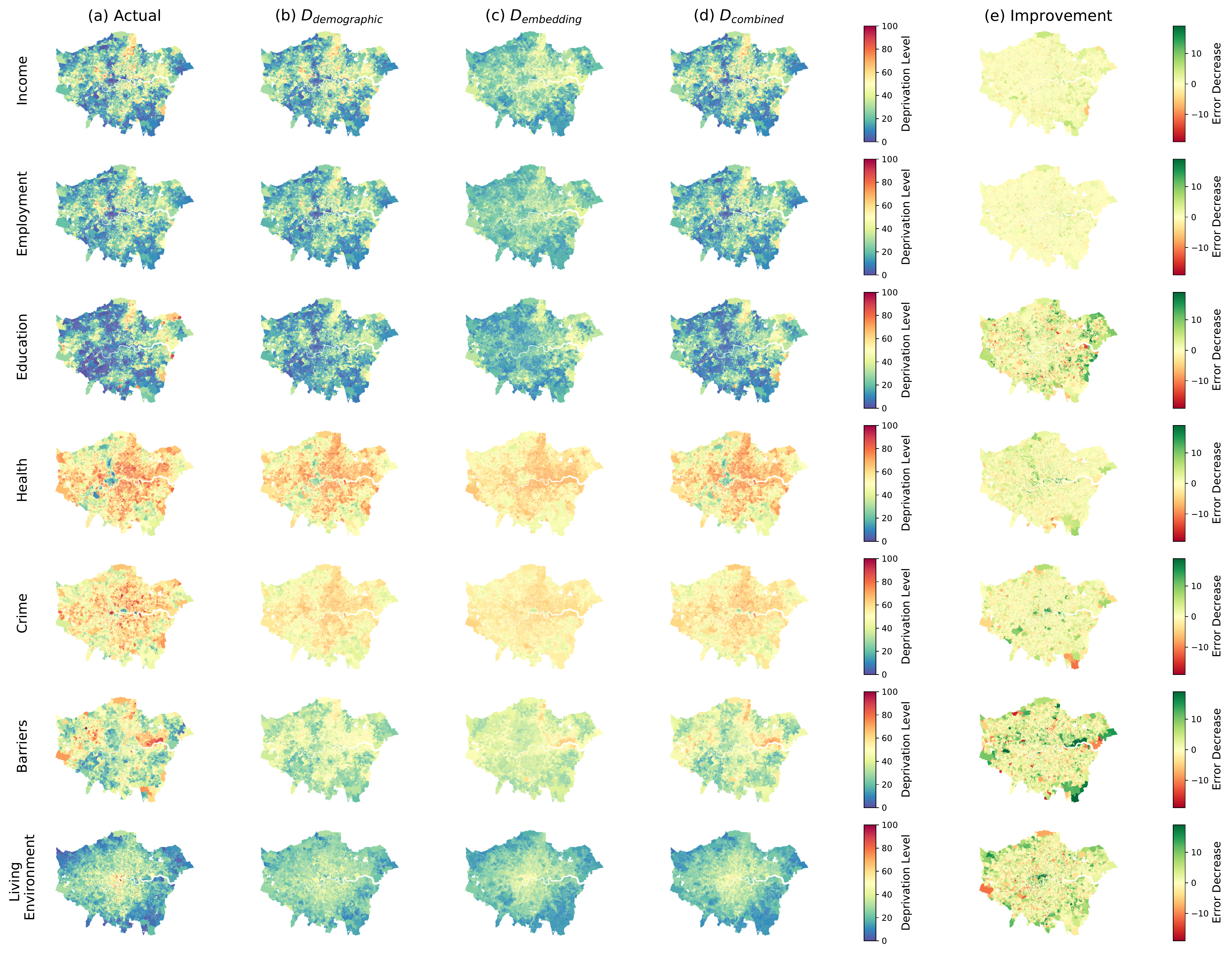}}
  \caption{LSOA level predictions. This figure shows the actual deprivation levels for all seven indices (a) against the predicted values, for the three subsets $D_{demographic}$ (a), $D_{embedding}$ (b) and $D_{combined}$ (c). This results set uses the GBM model, and for the $D_{embedding}$ and $D_{combined}$ results, the SimCLR embeddings are adopted using the largest embedding size (512) with mean-pooling. The improvement plots (e) show the geographical distribution of the error improvement in the $D_{combined}$ subset compared to the $D_{demographic}$ subset.}
  \label{fig:GeoError_full}
\end{figure}
\end{document}